%% file: 0_main.tex
\documentclass[letterpaper, 10 pt, journal, twocolumn]{Config/IEEEtran}
\usepackage[utf8]{inputenc}
\usepackage{bm}
\usepackage{amsmath}
\usepackage{amssymb}
\usepackage{mathtools}
\usepackage{graphicx}
\usepackage{subcaption}
\usepackage{stfloats}

\IEEEoverridecommandlockouts

\title{The Use of Implicit Human Motor Behaviour in the Online Personalisation of Prosthetic Interfaces}
\author{Ricardo Garcia-Rosas, Tianshi Yu, Denny Oetomo, Chris Manzie, Ying Tan, and Peter Choong \thanks{This project is funded by the Valma Angliss Trust.} \thanks{R. Garcia-Rosas, T. Yu, D. Oetomo, C. Manzie, and Y. Tan are with the School of Electrical, Mechanical and Infrastructure Engineering, and P. Choong with the Department of Surgery, The University of Melbourne, VIC 3010, Australia. {\tt \{ricardog,tianshiy\}@student.unimelb.edu.au; \{doetomo,manziec,yingt,pchoong\}@unimelb.edu.au}.}}

\begin{document}

\maketitle

\begin{abstract}
In \cite{Garcia-Rosas2019}, the authors proposed a data-driven optimisation algorithm for the personalisation of human-prosthetic interfaces, demonstrating the possibility of adapting prosthesis behaviour to its user while the user performs tasks with it. This method requires that the human and the prosthesis personalisation algorithm have same pre-defined objective function. This was previously ensured by providing the human with explicit feedback on what the objective function is. However, constantly displaying this information to the prosthesis user is impractical. Moreover, the method utilised task information in the objective function which may not be available from the wearable sensors typically used in prosthetic applications. In this work, the previous approach is extended to use a prosthesis objective function based on implicit human motor behaviour, which represents able-bodied human motor control and is measureable using wearable sensors. The approach is tested in a hardware implementation of the personalisation algorithm on a prosthetic elbow, where the prosthetic objective function is a function of upper-body compensation, and is measured using wearable IMUs. Experimental results on able-bodied subjects using a supernumerary prosthetic elbow mounted on an elbow orthosis suggest that it is possible to use a prosthesis objective function which is implicit in human behaviour to achieve collaboration without providing explicit feedback to the human, motivating further studies.
\end{abstract}


%\input{Outlines/outline_v1.tex}
\input{1_introduction_short.tex}
\input{2_system_short.tex}
\input{3_objectiveFunction.tex}
\input{4_personalisationExperiments.tex}
\input{5_discussion.tex}
\input{6_conclusion.tex}

%% ----------------------------------------------------------------
\bibliographystyle{ieeetr}  % Use the "unsrtnat" BibTeX style for formatting the Bibliography
\bibliography{library}  % The references (bibliography) information are stored in the file named "Bibliography.bib"

\end{document}

%% file: 1_introduction_short.tex
%
% Intro
%
\section{Introduction}
%\textbf{What is the motivation of extending our previous work? - The challenges introduced by the practical application.}

Personalisation of human-prosthetic interfaces has been shown to be a necessary step to their successful implementation \cite{Merad2018, Garcia-Rosas2018EMBC,  Biddiss2007a}. The authors previously proposed a method to perform the personalisation process autonomously while the prosthesis user performs a task with the prosthetic device \cite{Garcia-Rosas2019}. This approach utilises data gathered online to evaluate a pre-defined measure of performance (objective function), which the personalisation algorithm seeks to optimise. 

The measure of performance used for personalisation defines the prosthetic's objective, while the human has its own internal measure of performance and thus objective. In collaborative human-robot applications, such as motion-based synergistic prosthesis \cite{Merad2016, Alshammary2018, Akhtar2017}, it is required that these objectives align, i.e. that both the human and the robot have the same objective functions \cite{Garcia-Rosas2019, Li2016}. In \cite{Garcia-Rosas2019}, this was ensured by using task information (reach accuracy and time) as the objective function in the personalisation algorithm and explicitly presenting it to the user. This is referred to as the ``explicit'' objective function henceforth. However, in practice this may not be possible as the data obtained by the prosthesis is given by wearable sensors, and task information may not be available. Moreover, it may neither be possible nor desirable to constantly display performance information to the user.

A possibility to relax the need of having the same measure of performance between the user and prosthesis is to use knowledge of the implicit motor behaviour of humans. Implicit motor behaviour refers to the processes that occur subconsciously in humans when performing physical tasks. This work hypothesises that by utilising a measure of performance that is theorised to be implicit in human motor behaviour, no explicit objective function will need to be provided to the prosthesis user to achieve human-prosthesis collaboration. %Thus the personalisation algorithm would be able to unobtrusively determine the best synergy for a given individual during normal prosthesis use.

%\textbf{This paper...}

%This work presents an extension of the work presented in \cite{Garcia-Rosas2019} which tests the hypothesis of using an implicit measure of performance to unobtrusively drive prosthesis personalisation. The proposed measure of implicit human motor performance is based on upper-body compensation motion, which is determined by data gathered from two wearable IMUs. The hypothesis is studied on able-bodied subjects using a supernumerary prosthetic elbow mounted on an elbow orthosis. Results show that using a surrogate for implicit human motor behaviour as a measure of performance can successfully drive the personalisation of human-prosthetic interfaces.

%% file: 2_system_short.tex
%
% System
%
%\section{Human-Prosthesis System with Adaptive Synergistic HPI}

\begin{figure}[hbt]
    \centering
    \includegraphics[width=\columnwidth]{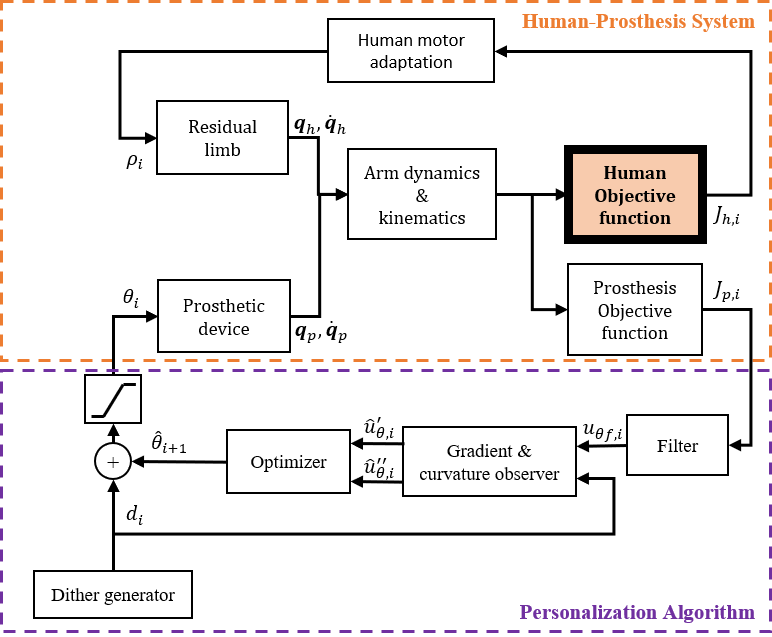}
    \caption{Human-prosthesis system with adaptive synergies. This work focuses on $J_h$.}%Both the human and prosthesis have their own internal objective functions that are being optimised; however, these may not be the same.}
    \label{fig:fullSystemPersonalisation}
\end{figure}

A block diagram of the human-prosthesis system and the personalisation algorithm is shown in Figure \ref{fig:fullSystemPersonalisation}. The human (residual) limb and prosthetic device interact to achieve a resultant hand motion, where the motion of the prosthetic device is a function of the synergy value and the human limb motion (time domain). As a result of human motor learning, humans change their motor behaviour over repeated use of the prosthetic device in order to use the device effectively according to an internal measure of performance, $J_h$, (iteration domain). There are currently multiple propositions for the internal composition of internal optimisation metrics used to generate optimal motor behaviour \cite{Kistemaker2014, Huang2012, Todorov2004}. This human-internal measure of performance is referred to as the ``implicit'' objective function henceforth.

On the other hand, the prosthesis objective function ($J_p$) used in the synergy personalisation algorithm is explicitly defined and chosen to achieve an objective with the prosthesis. Due to the recent development of such adaptive system in \cite{Garcia-Rosas2019}, the effects of the choice and design of the objective function on the resultant human motor behaviour are unknown. However, in order to design such objective functions, a set of conditions must be met. This is described in a model of the synergy-to-performance behaviour inherent in the human-prosthesis system \cite{Garcia-Rosas2019}. In summary, the objective function must be designed such that there exists a synergy ($\theta^*$) that achieves the best performance ($J_p^*$), and human motor learning must be reflected in the performance ($J_p$) as monotonically increasing until reaching a steady-state.

To achieve collaboration, and thus convergence of the personalisation algorithm, the human and prosthetic objective functions must be as close as possible ($J_p \rightarrow J_h$) \cite{Garcia-Rosas2019, Li2016}. Here, the use of a measure of implicit human motor behaviour in the prosthesis objective function is proposed. The hypothesis is that the use of a prosthesis objective function that acts as a surrogate for the internal human objective function will allow the personalisation process to take place without the need to provide explicit feedback to the human. The hypothesis is tested by using upper-body compensation motion as a surrogate for human effort while performing a reaching task, which is determined by data gathered from two wearable IMUs. The hypothesis is studied on able-bodied subjects using a supernumerary prosthetic elbow mounted on an elbow orthosis. Results show that using a surrogate for implicit human motor behaviour as a measure of performance can successfully drive the personalisation of human-prosthetic interfaces.

%% file: 3_objectiveFunction.tex
%
% Cost function
%
\section{Upper-body Compensation as a Measure of Performance for Prosthesis Personalisation}

%\textbf{What is our approach to the design of the measure of performance and why.}

This section presents the methodology followed to determine an IMU-based measure of performance for prosthetic reaching. The choice of using compensation motion is based on its use as a clinical measure of prosthetic performance \cite{Carey2008, Metzger2012}. Compensation motion can be measured with body-mounted IMU sensors \cite{Merad2018}, making it a good candidate for a wearable sensor-based measure of performance in prosthetics. Moreover, upper-body compensation motion is used by prosthesis users to overcome the limitations of their prosthetic devices. It is also observed in a wide range of ADLs, and trunk and shoulder motion are the natural range extenders of reach in able-bodied motion \cite{Carey2008, Metzger2012}. 

The following steps were followed to develop the objective function: 1) the relationship between the prosthesis synergy and upper-body motion was experimentally evaluated to determine the relevant components to be used in the design of the objective function (Section \ref{sec:evaluation}). 2) An objective function was designed such that the necessary conditions for its use in the personalisation algorithm are satisfied (Section \ref{sec:objectiveFunction}).

\subsection{Evaluating the Relationship Between Synergy and Upper-body Motion}
\label{sec:evaluation}
%\textbf{How do we evaluate the relationship? aka experimental methodology.}

\begin{figure*}[htb]
    \centering
    \begin{subfigure}[t]{0.32\textwidth}
        \centering
        \includegraphics[width=\textwidth]{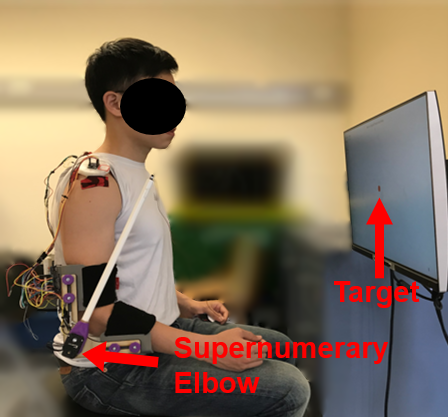}
        \caption{Starting position for the reaching task, the target on the screen, and the supernumerary elbow.}
        \label{fig:startAndTarget}
    \end{subfigure}
    ~
    \begin{subfigure}[t]{0.32\textwidth}
        \centering
        \includegraphics[width=\textwidth]{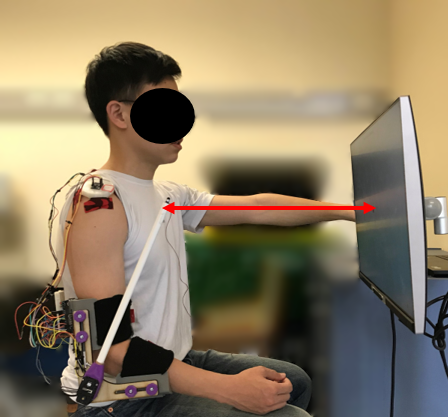}
        \caption{The screen location was adjusted to each subject's arm length and height.}
        \label{fig:screenLocation}
    \end{subfigure}
    ~
    \begin{subfigure}[t]{0.32\textwidth}
        \centering
        \includegraphics[width=1.075\textwidth]{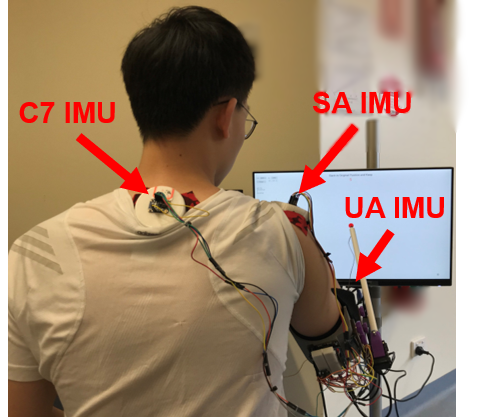}
        \caption{Sensors were placed on the C7 vertebrae (C7), the shoulder acromion (SA), upper-arm (UA), and lower arm (able-bodied only).}
        \label{fig:sensorPlacement}
    \end{subfigure}
    \caption{Experimental set-up used for determining the relationship between synergy and upper-body motion, and to demonstrate the proposed measure of performance.}
    \label{fig:experimentSetup}
\end{figure*}

To determine the relationship between the synergy and upper-body compensation motion an experiment with three able-bodied subjects wearing a supernumerary prosthetic elbow mounted on an elbow orthosis (Figure \ref{fig:startAndTarget}) was performed. 

\subsubsection{Hardware set-up} Bosch BNO055 IMUs were mounted on the subject's C7 vertebrae to measure trunk motion (C7), on the shoulder acromion (SA) to measure shoulder displacement, on the upper-arm (UA) to be used for the synergy, and on the lower-arm (LA) to get able-bodied elbow motion data. Sensor placement is shown in Figure \ref{fig:sensorPlacement}. The C7 and SA sensors were used to determine the trunk and shoulder forward displacement respectively. Displacement was calculated using the subject's body measurements, trunk length and C7 to shoulder acromion distance, and the estimated joint angle from the IMUs. The upper-body was considered to be a set of rigid links. Data gathering was done using an Arduino M0 Zero and an application developed in Visual Studio/C\#.

\subsubsection{Task description} The task required subjects to reach forward from a neutral seating position and touch a target on a screen. The starting position and target are shown in Figure \ref{fig:startAndTarget}. The screen location was adjusted for each subject to be within their able-bodied reaching distance. The screen position was set by using the subject's arm as a reference while the arm was held straight forward as shown in Figure \ref{fig:screenLocation}. The forward distance and height of the screen were set at the position of the subject's wrist joint, while the lateral position was set such that the centre of the screen was aligned with the centre of the subject's chest.

\subsubsection{Protocol description} Subjects were first asked to perform the task with their arm (able-bodied) for 30 iterations to obtain their able-bodied motor behaviour as benchmark. This is referred to as the able-bodied case henceforth. Then the supernumerary synergistic prosthetic elbow was mounted on their arm using an elbow orthosis and were provided with sufficient training with the device in order to minimise the effects of motor learning on the synergy-motion results. This is referred to as the prosthetic case henceforth. Subjects repeated the reaching task for 200 iterations, with the synergy value changing every 5 iterations ($\Delta\theta = 0.05$). the synergy used was $\dot{q}_e = \theta \dot{q}_s$, where $\dot{q}_e$ is the prosthetic elbow extension, $\dot{q}_s$ the shoulder extension, and $\theta\in[0.8, 2.7]$ the synergy parameter. The procedure was approved by the University of Melbourne Human Research Ethics Committee, project number 1750711.2. Informed consent was received from all subjects in the study. Results from experiments in this manuscript can be found in https://git.io/JvRZs.

\subsubsection{Synergy vs. Trunk compensation} 
\begin{figure*}[htb]
    \centering
    \begin{subfigure}[t]{0.32\textwidth}
        \centering
        \includegraphics[width=1.075\textwidth]{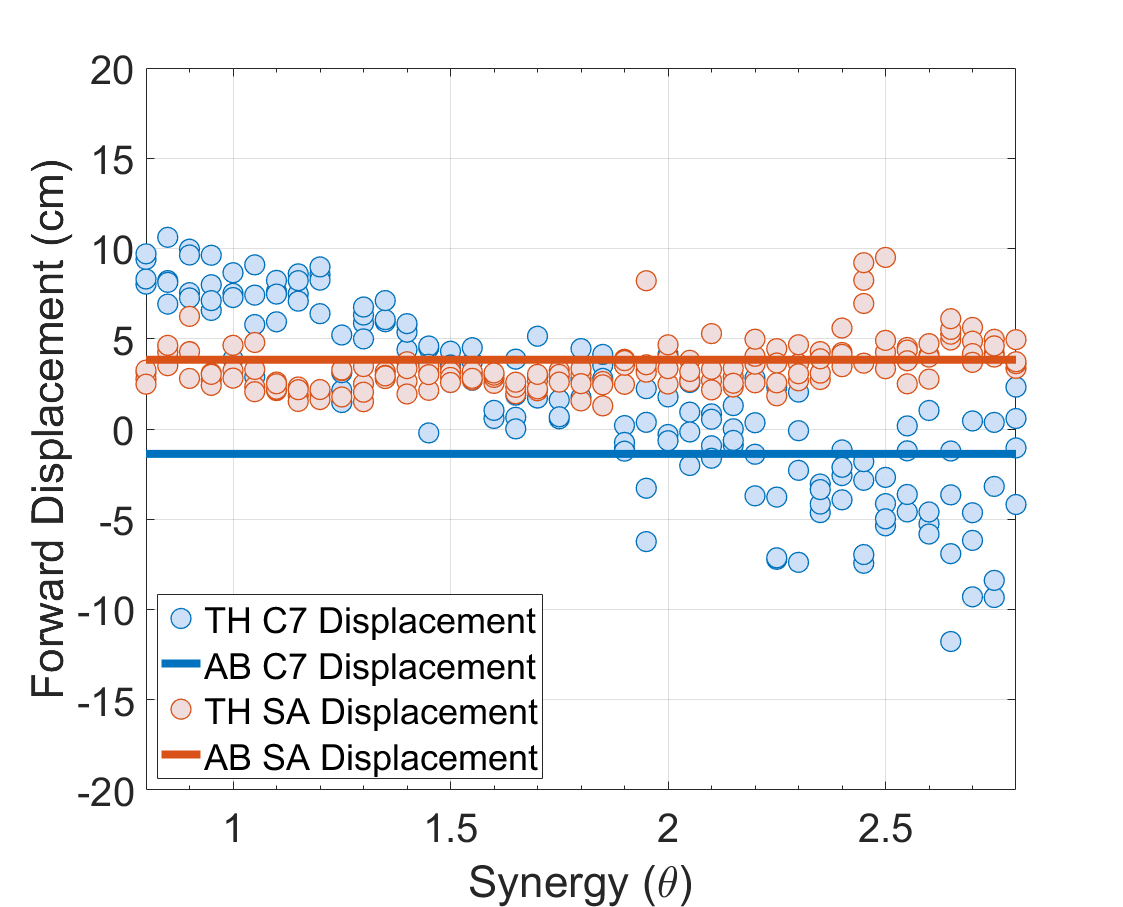}
        \caption{Subject 1.}
        \label{fig:sweepS1}
    \end{subfigure}
    ~
    \begin{subfigure}[t]{0.32\textwidth}
        \centering
        \includegraphics[width=1.075\textwidth]{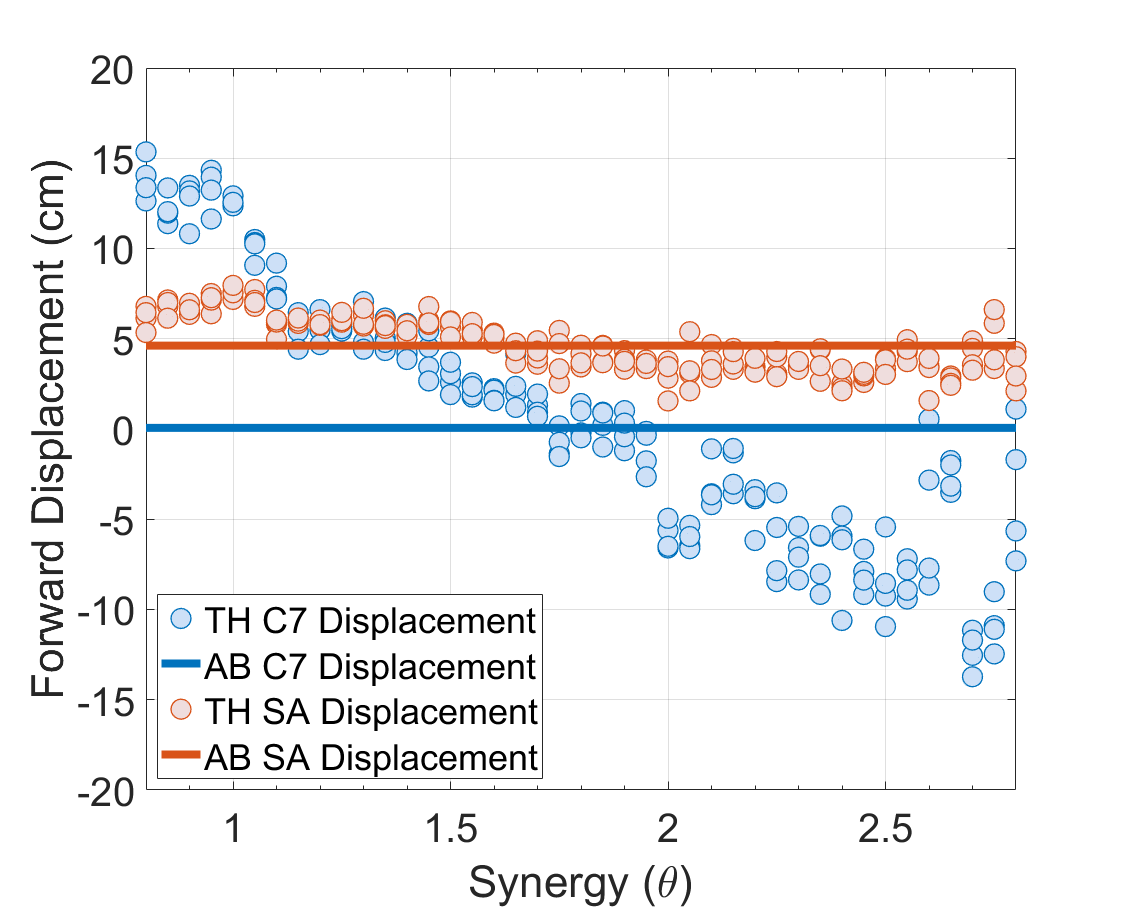}
        \caption{Subject 2.}
        \label{fig:sweepS2}
    \end{subfigure}
    ~
    \begin{subfigure}[t]{0.32\textwidth}
        \centering
        \includegraphics[width=1.075\textwidth]{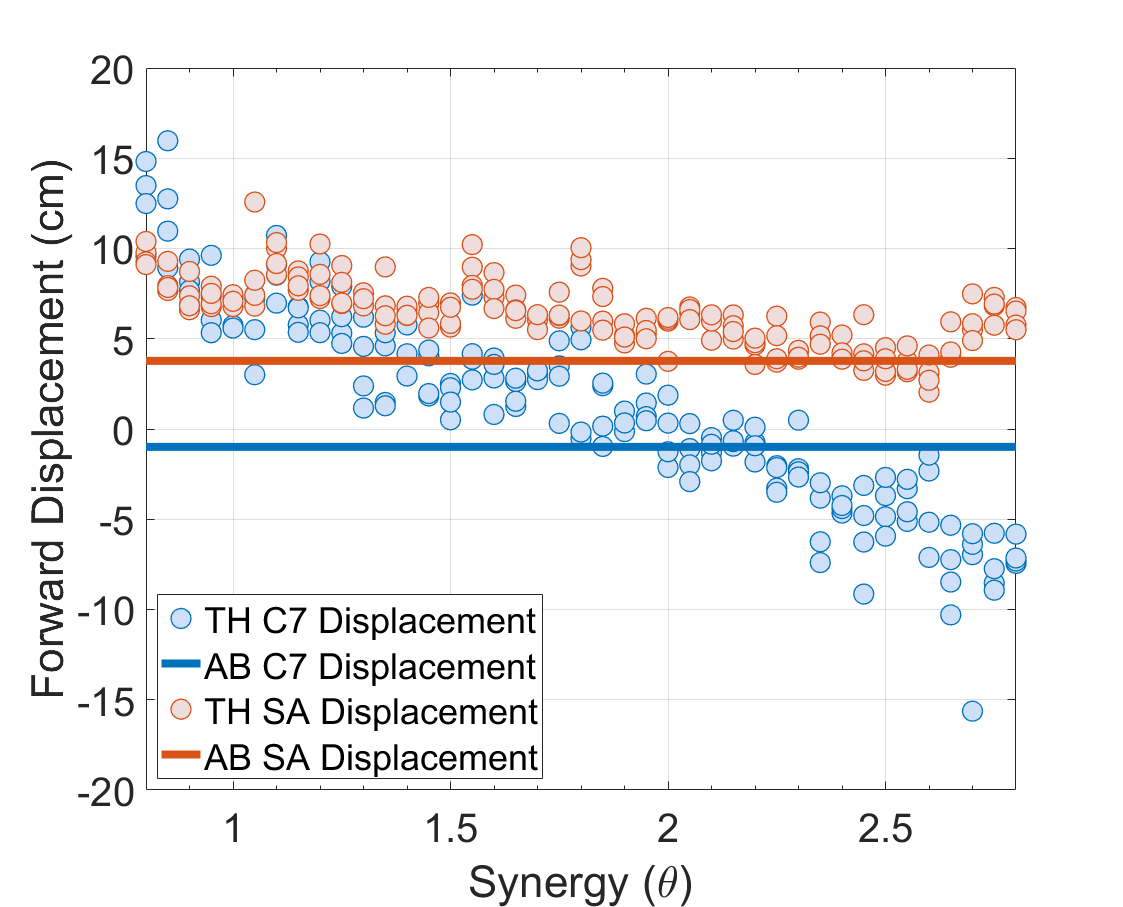}
        \caption{Subject 3.}
        \label{fig:sweepS3}
    \end{subfigure}
    
    % \begin{subfigure}[t]{0.32\textwidth}
    %     \centering
    %     \includegraphics[width=1.075\textwidth]{Figures/Results/HPIs/8-hpi-bu-1.eps}
    %     \caption{Subject 1.}
    %     \label{fig:C7UWD_1}
    % \end{subfigure}
    % ~
    % \begin{subfigure}[t]{0.32\textwidth}
    %     \centering
    %     \includegraphics[width=1.075\textwidth]{Figures/Results/HPIs/8-hpi-bu-2.eps}
    %     \caption{Subject 2.}
    %     \label{fig:C7UWD_2}
    % \end{subfigure}
    % ~
    % \begin{subfigure}[t]{0.32\textwidth}
    %     \centering
    %     \includegraphics[width=1.075\textwidth]{Figures/Results/HPIs/8-hpi-bu-3.eps}
    %     \caption{Subject 3.}
    %     \label{fig:C7UWD_3}
    % \end{subfigure}
    \caption{Relationship between trunk (C7) and shoulder (SA) forward displacement, and synergy ($\theta$) for three subjects. The x-axis presents the synergy value while the y-axis the trunk and shoulder forward displacement. Blue and red circles represent C7 and SA displacements for the prosthetic case, respectively. Blue and red lines represent mean able-bodied C7 and SA displacements, respectively.}
    \label{fig:synergyisplacementMap}
\end{figure*}

Experimental results for trunk forward compensation for the three subjects are presented in Figure \ref{fig:synergyisplacementMap}. The blue line represents average trunk displacement for the able-bodied case while the blue circles for the prosthetics case. It can be observed that trunk displacement for the able-bodied case is close to zero for the given reaching task. In the prosthetics case, trunk displacement has a near linear relationship with the synergy. Moreover, it can be seen that the prosthetic case data intersects the subject's able-bodied displacement line. This intersection represents the subject's optimal synergy value which achieves able-bodied-like trunk displacement. Intuitively, this means that as the synergy value moves away from the optimal, the individual will recruit trunk motion to compensate for the elbow not extending enough, or over-extending. This behaviour is highly desirable for personalisation purposes as it suggests that there is a unique synergy that minimises trunk displacement. Moreover, the able-bodied line intersection is at a different synergy value across subjects and the slope of the synergy-displacement map differs across subjects, highlighting individuality in motor behaviour. 

% Results

\subsubsection{Synergy vs. Shoulder compensation} Experimental results for shoulder forward compensation for the three subjects are presented in Figure \ref{fig:synergyisplacementMap}. The red line represents average able-bodied shoulder displacement while the red circles the prosthetics case. From these results, it can be seen that shoulder forward displacement is always present in the reaching motion, with able-bodied displacement of about four centimetres for the given task. In the prosthetics case, different compensation strategies can be observed across subjects in the shape of the synergy-displacement map. This highlights individuality and preference in motor behaviour. Similarly to trunk compensation, a crossing of the synergy-displacement map with the able-bodied average displacement is present, which in this case is at $4$cm.

\subsection{A Compensation Motion-based Objective Function for Prosthesis Personalisation}
\label{sec:objectiveFunction}
The purpose of the design of the objective function is to drive the behaviour of the personalisation algorithm. In this work, this means minimising compensation motion. Therefore, the resultant Synergy-Compensation map needs to satisfy the convexity assumption in \cite[Assumption 1]{Garcia-Rosas2019}. Given that the synergy-displacement relationship observed in Figure \ref{fig:synergyisplacementMap} is nearly linear, a candidate objective function that satisfies this assumption is the convex combination of quadratic terms. For compensation motion, this is given by:
\begin{equation}
\label{eq:costFunction}
    J_p = \alpha (\bar{x}_{t} - x_{t})^{2} + (1-\alpha) (\bar{x}_{s} - x_{s})^2,
\end{equation}
where $0 < \alpha < 1$ is the weight to be determined, $x_t$ the trunk forward displacement, $\bar{x}_{t}$ the desired trunk forward displacement (able-body-like), $x_s$ the shoulder forward displacement, and $\bar{x}_{s}$ the desired shoulder forward displacement (able-body-like). Due to the minimisation objective, $J_p$ will be referred as cost henceforth.

A rigorous choice of the weight $\alpha$ would require determining the involvement of each joint in the reaching motion, and thus an analysis of human motor behaviour and the theorised internal optimisation mechanisms for human motion planning \cite{Kistemaker2014, Anderson2001}. This is out of the scope of this paper and will be investigated in future work. Therefore, it was chosen to equally minimise both trunk and shoulder displacement by setting $\alpha = 0.5$, and focus on determining the viability of the proposed method from a practical perspective. 

\begin{figure*}[htb]
    \centering
    \begin{subfigure}[t]{0.32\textwidth}
        \centering
        \includegraphics[width=1.075\textwidth]{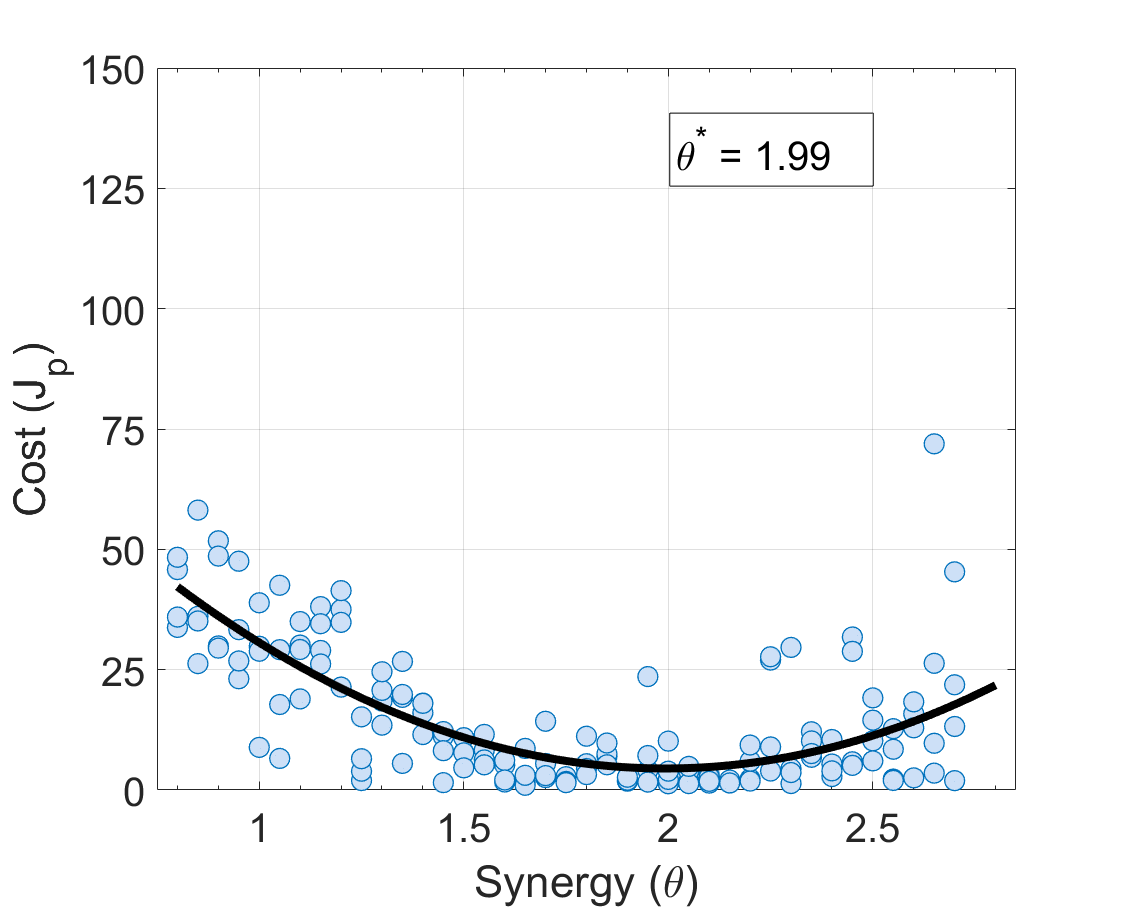}
        \caption{Subject 1.}
        \label{fig:map_1}
    \end{subfigure}
    ~
    \begin{subfigure}[t]{0.32\textwidth}
        \centering
        \includegraphics[width=1.075\textwidth]{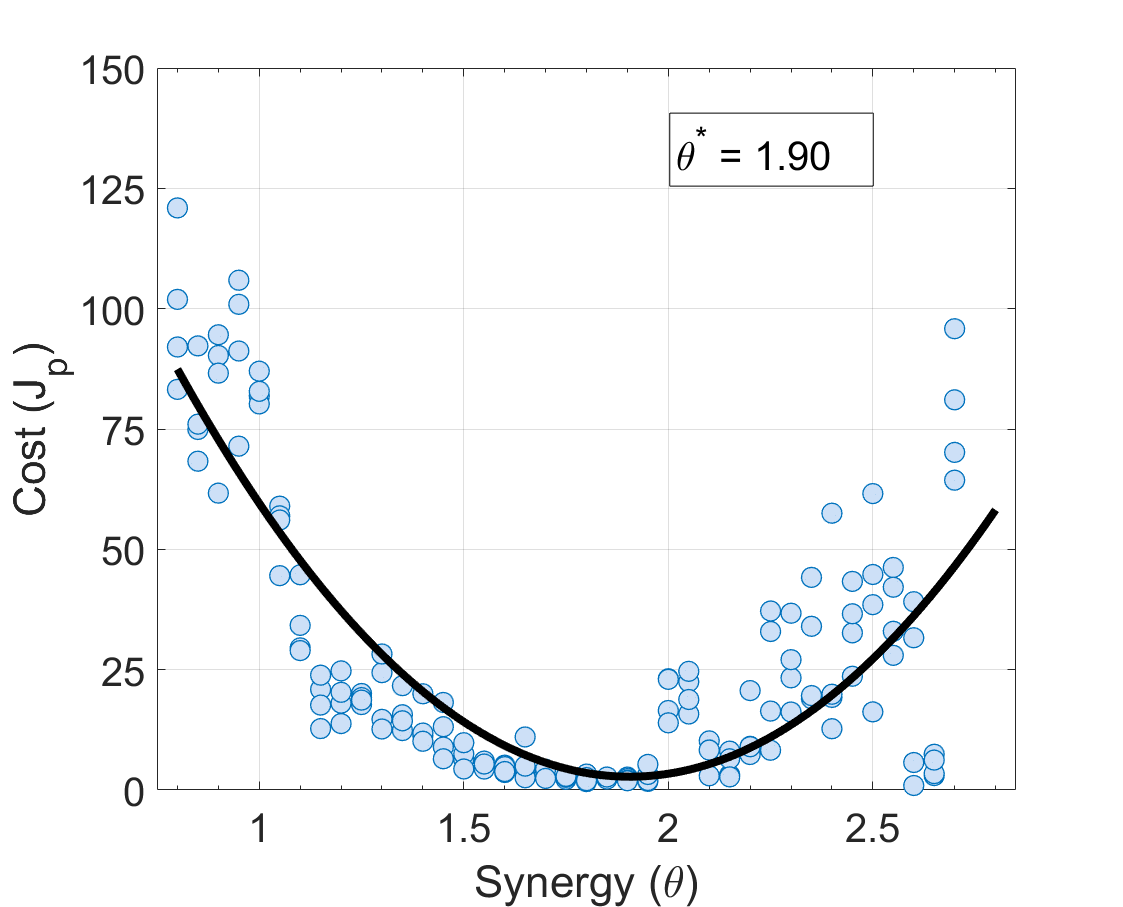}
        \caption{Subject 2.}
        \label{fig:map_2}
    \end{subfigure}
    ~
    \begin{subfigure}[t]{0.32\textwidth}
        \centering
        \includegraphics[width=1.075\textwidth]{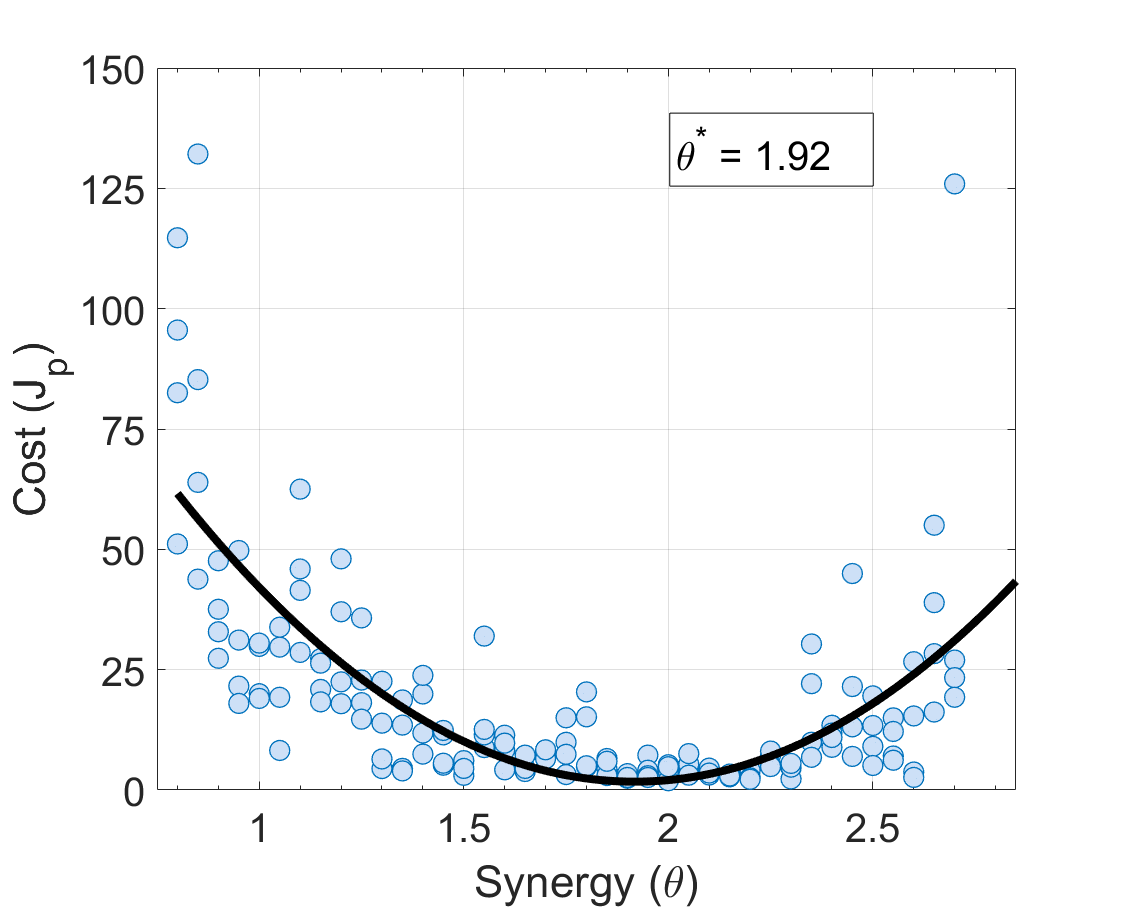}
        \caption{Subject 3.}
        \label{fig:map_3}
    \end{subfigure}
    \caption{Synergy-Performance map ($J_p(\theta)$) for three subjects. The x-axis presents the synergy value while the y-axis the performance as given by eqn. \ref{eq:costFunction}. The experimental results are shown in blue while the fitted quadratic map is shown in black. The optimal synergy ($\theta^*$) for each subject, as given by the fitted quadratic map, is shown in each plot.}
    \label{fig:synergyPerformanceMaps}
\end{figure*}

The obtained synergy-cost maps ($J_p(\theta)$) for the three subjects are presented in Figure \ref{fig:synergyPerformanceMaps}. The experimental data is represented by the blue circles, while the quadratic polynomial fit to this data is shown by the black lines. The estimated optimal synergies ($\theta^*$), given by the polynomial, are 1.99, 1.90, and 1.92, for each respective subject. However, it is important to note from the experimental data that the minimum cost is observed for a range of synergy values. As expected the synergy-cost maps ($J_p(\theta)$) show desirable features for online personalisation. These results shows that the proposed compensation motion-based objective function satisfies the condition for the implementation of the algorithm presented in \cite{Garcia-Rosas2019}.

%% file: 4_personalisationExperiments.tex
%
% Personalisation experiments
%
\section{Compensation Motion-based Prosthesis Personalisation Experiments}
An experiment using the same set-up presented in Section \ref{sec:evaluation} was performed to evaluate the behaviour of the personalisation algorithm with the proposed measure of performance. The experiment involved nine able-bodied subject, the same three subjects in Section \ref{sec:evaluation} and six additional subjects that were completely new to the set-up and task. These experiments were performed three months after the original sweep experiments to minimise the bias introduced by motor learning on the results of the first three subjects. The same task as in \ref{sec:evaluation} was performed by the subjects; however, in this case, the personalisation algorithm was used to iteratively adjust the synergy parameter ($\theta$) instead of performing a synergy value sweep. The following algorithm tuning parameters were used: $\omega_o = \pi/4$, $a = 0.06$, $k = 0.0008$, $\epsilon = 0.1$, and $L = \begin{bmatrix} 0.3840 & 0.6067 & -0.2273 & -0.8977 & -1.0302 \end{bmatrix}^T$. Detailed information on these parameters can be found in \cite{Garcia-Rosas2019}.

Subjects were not informed to what was being evaluated by the prosthesis and were only indicated to perform the reaching task with the supernumerary prosthetic elbow. Subjects performed the task for 80 iterations, with a one minute rest after 40 iterations. The target displacements trunk and shoulder displacements, $\bar{x}_{t}$ and $\bar{x}_{s}$, were set to zero. The choice of zero trunk displacement is based on able-bodied trunk displacement. The choice of zero shoulder displacement was to determine whether the implicit prosthetic objective function could influence ``natural'' human motion.

\begin{figure*}[htb]
    \centering
    \begin{subfigure}[t]{0.28\textwidth}
        \centering
        \includegraphics[width=1.0\textwidth]{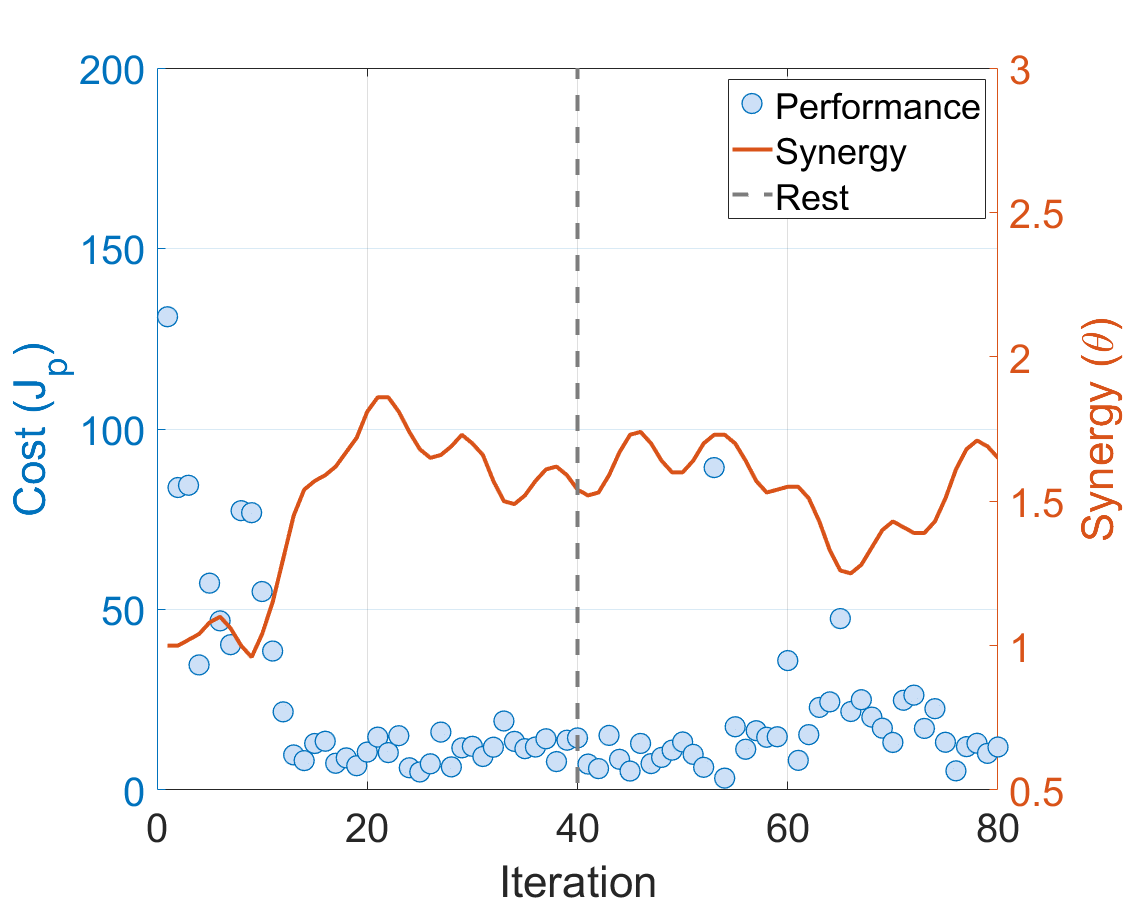}
        \caption{Subject 1.}
        \label{fig:thetaJS1}
    \end{subfigure}
    ~
    \begin{subfigure}[t]{0.28\textwidth}
        \centering
        \includegraphics[width=1.0\textwidth]{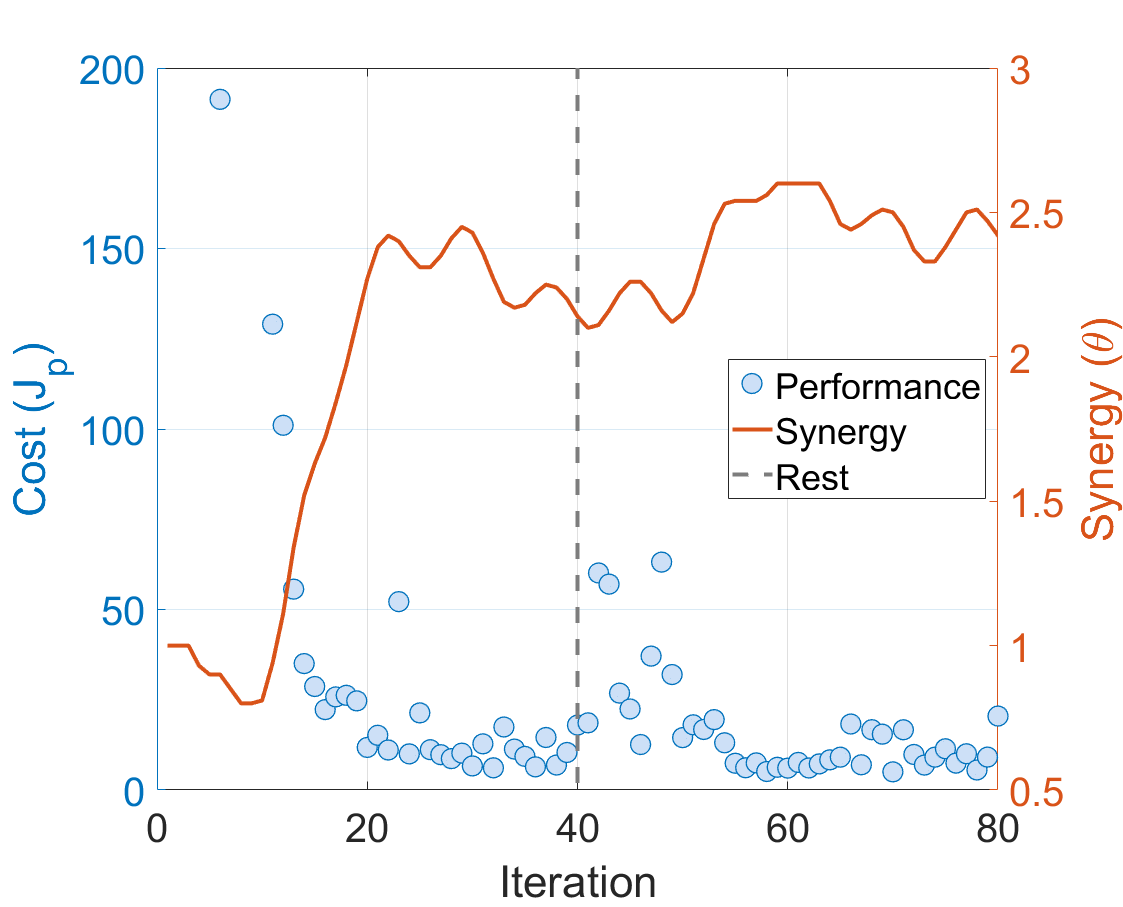}
        \caption{Subject 2.}
        \label{fig:thetaJS2}
    \end{subfigure}
    ~
    \begin{subfigure}[t]{0.28\textwidth}
        \centering
        \includegraphics[width=1.0\textwidth]{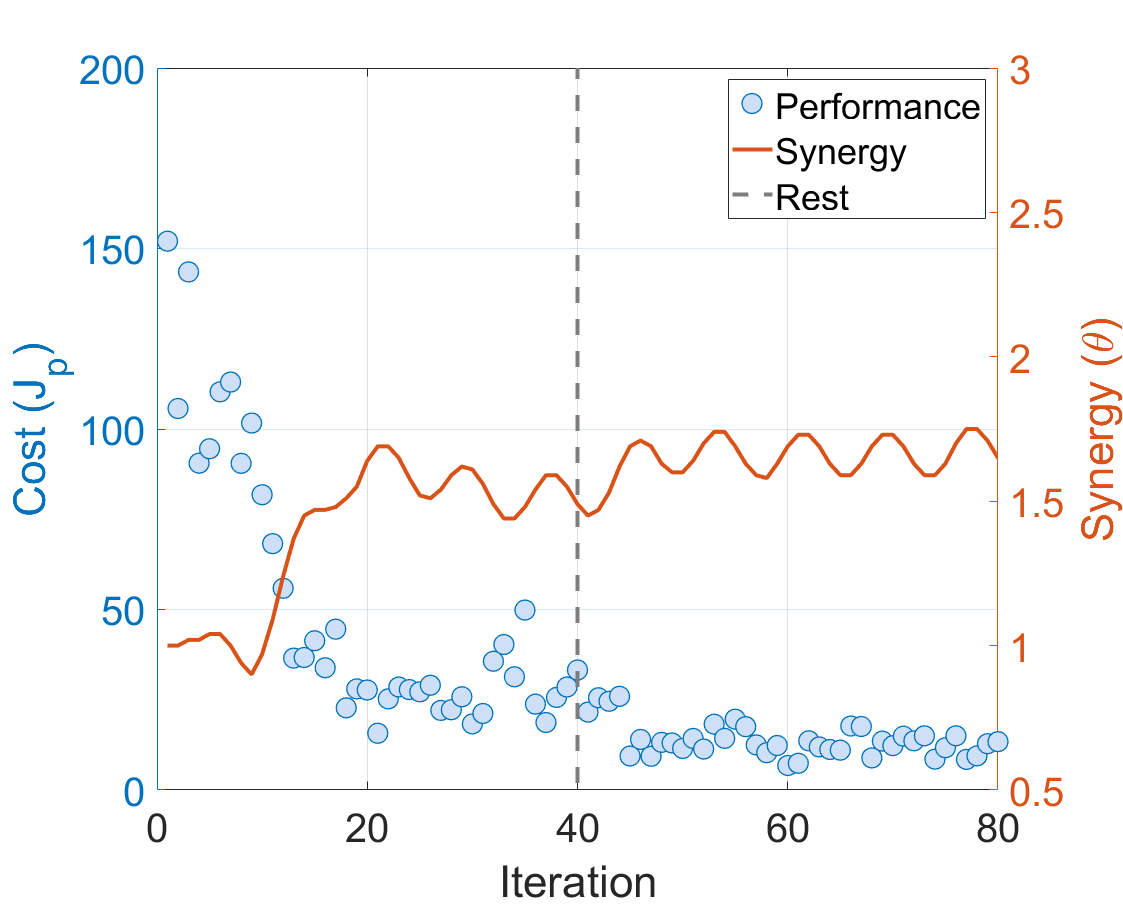}
        \caption{Subject 3.}
        \label{fig:thetaJS3}
    \end{subfigure}
    
    \begin{subfigure}[t]{0.28\textwidth}
        \centering
        \includegraphics[width=1.0\textwidth]{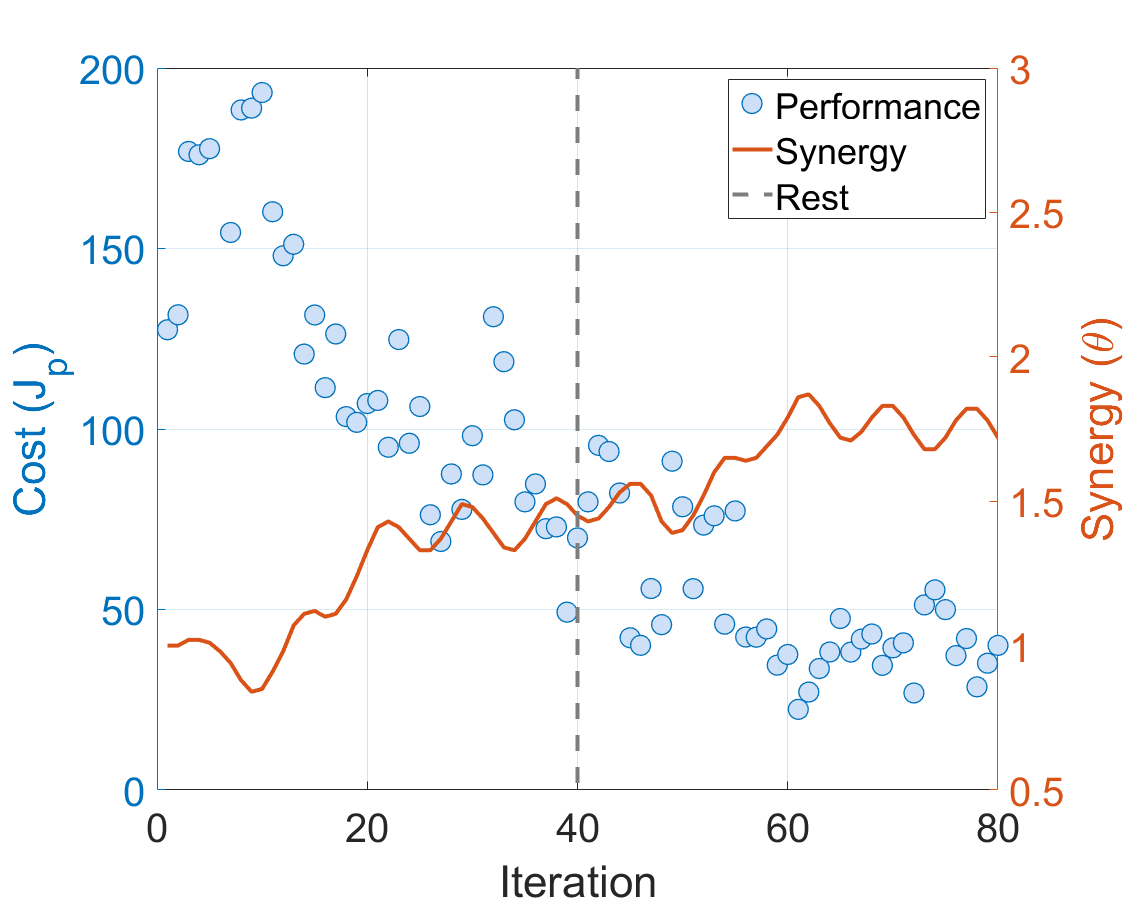}
        \caption{Subject 4.}
        \label{fig:thetaJS4}
    \end{subfigure}
    ~
    \begin{subfigure}[t]{0.28\textwidth}
        \centering
        \includegraphics[width=1.0\textwidth]{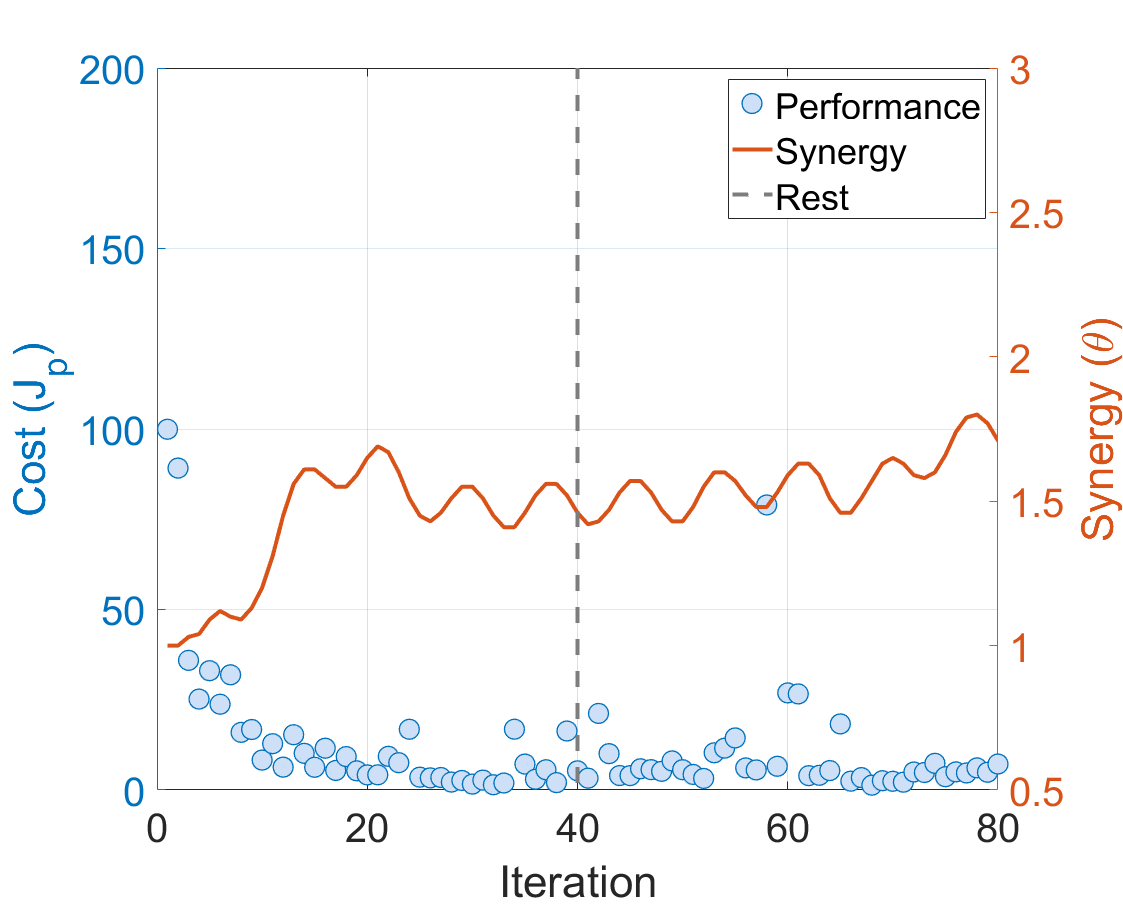}
        \caption{Subject 5.}
        \label{fig:thetaJS5}
    \end{subfigure}
    ~
    \begin{subfigure}[t]{0.28\textwidth}
        \centering
        \includegraphics[width=1.0\textwidth]{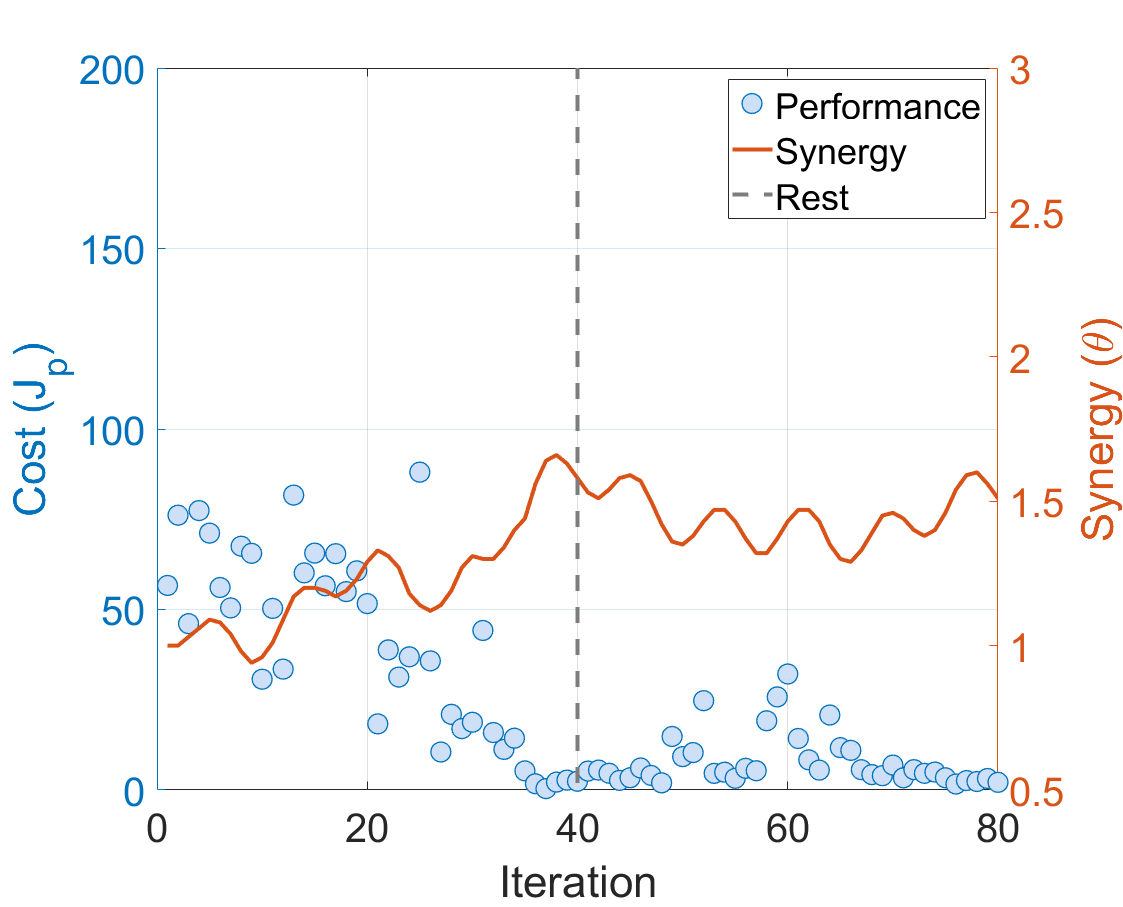}
        \caption{Subject 6.}
        \label{fig:thetaJS6}
    \end{subfigure}
    
    \begin{subfigure}[t]{0.28\textwidth}
        \centering
        \includegraphics[width=1.0\textwidth]{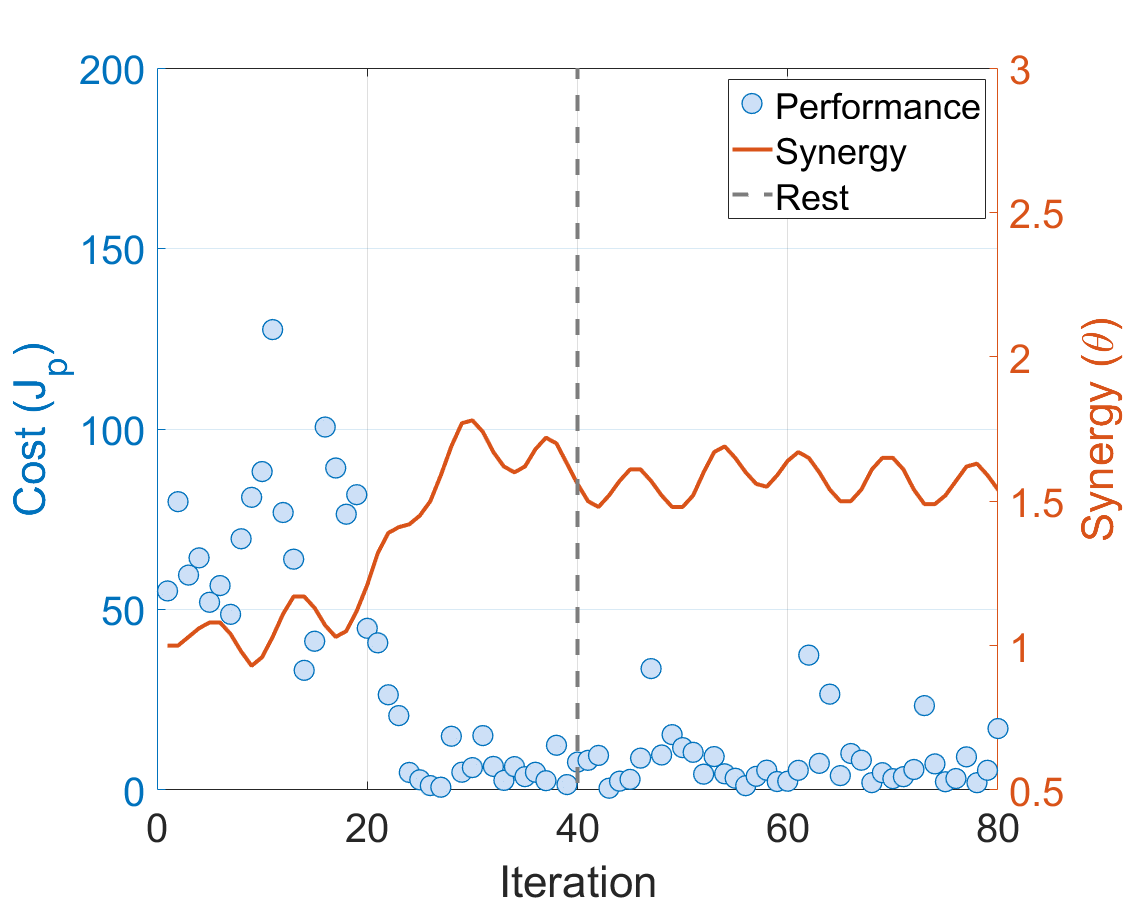}
        \caption{Subject 7.}
        \label{fig:thetaJS7}
    \end{subfigure}
    ~
    \begin{subfigure}[t]{0.28\textwidth}
        \centering
        \includegraphics[width=1.0\textwidth]{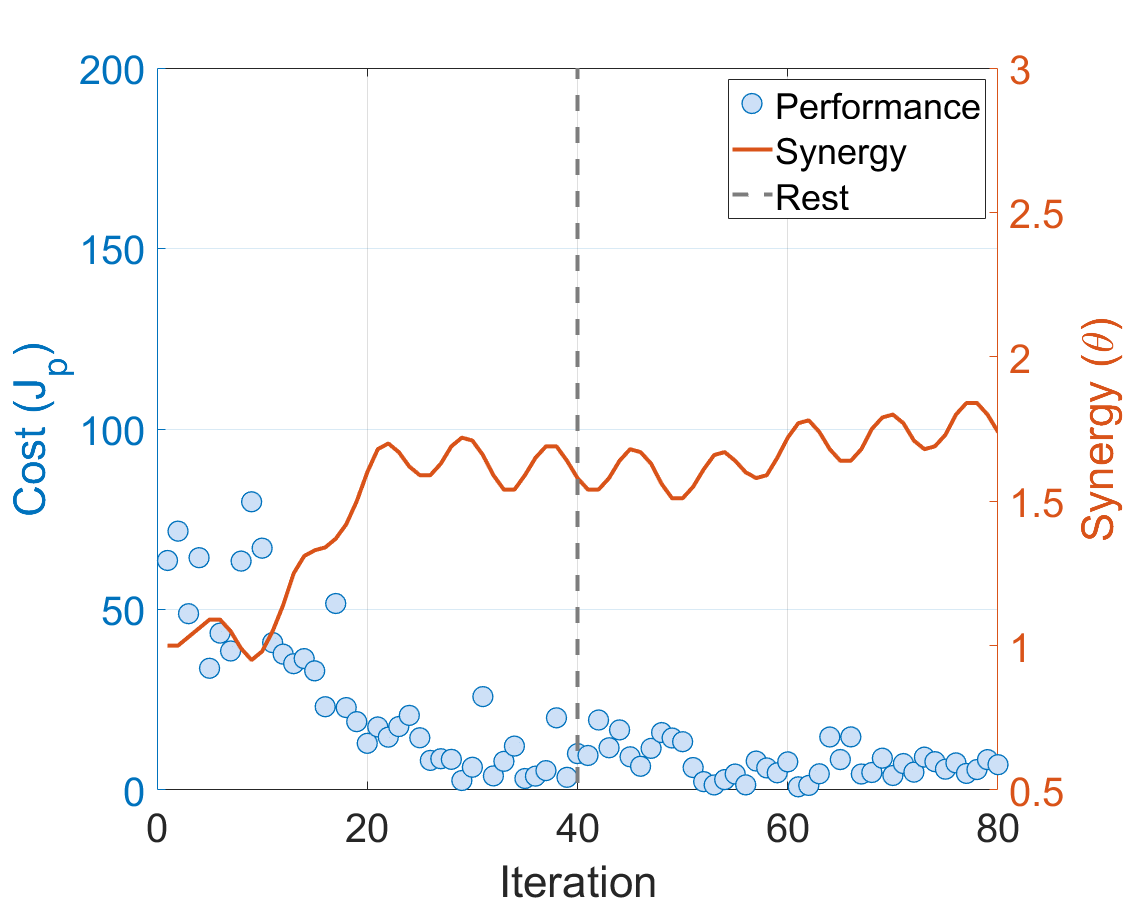}
        \caption{Subject 8.}
        \label{fig:thetaJS8}
    \end{subfigure}
    ~
    \begin{subfigure}[t]{0.28\textwidth}
        \centering
        \includegraphics[width=1.0\textwidth]{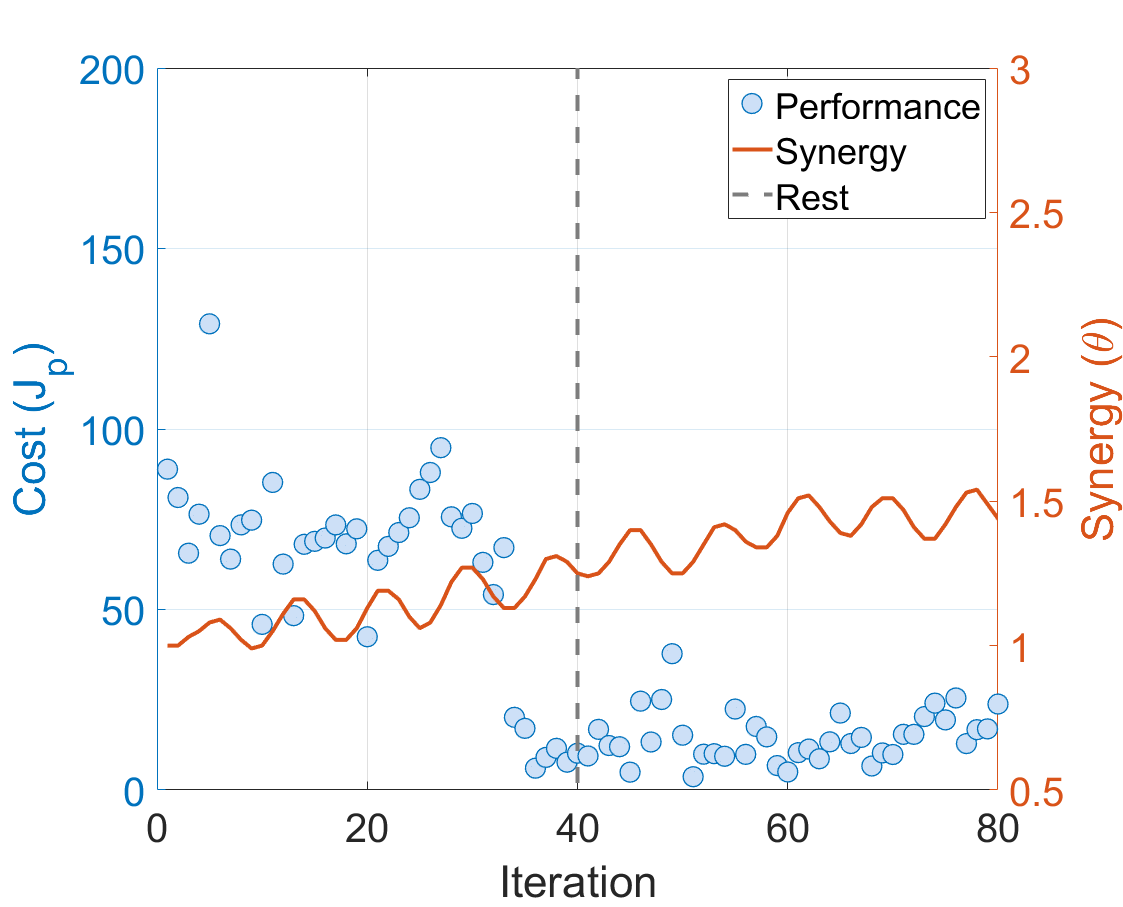}
        \caption{Subject 9.}
        \label{fig:thetaJS9}
    \end{subfigure}
    \caption{Synergy value ($\theta$) and cost ($J_p$) over iterations results for nine able-bodied subjects. The red line represents the synergy value ($\theta$) while the blue dots the cost ($J$) as defined by eqn. (\ref{eq:costFunction}). The grey dotted line represents when the subjects were given rest time.}
    \label{fig:synergyPerformanceResults}
\end{figure*}

Figure \ref{fig:synergyPerformanceResults} presents algorithm performance results. These results show the synergy value and cost, as defined in equation (\ref{eq:costFunction}), over the 80 iterations of the task. On average, the algorithm reached steady-state within 30 iterations of the task. With this given synergy steady-state ($\theta_{ss}$), subjects significantly minimised the cost ($J_p$), driving it close to zero, meaning that compensation motion was close to zero. This personalisation algorithm behaviour is comparable to the results presented in \cite{Garcia-Rosas2019}. However, it is important to highlight that these two approaches may not converge to the same optimal synergy. This is due to the different objective functions in use. There are two other significant observations. First, for subjects 1-3, the steady-state synergy converged to a synergy within the range that minimises the cost identified in the synergy-cost map in Figure \ref{fig:synergyPerformanceMaps}. Second, the motor behaviour of subjects 1 and 9 is irregular and is reflected in the performance of the algorithm. These will be discussed in the next section

\begin{figure*}[htb]
    \centering
    \begin{subfigure}[t]{0.28\textwidth}
        \centering
        \includegraphics[width=1.0\textwidth]{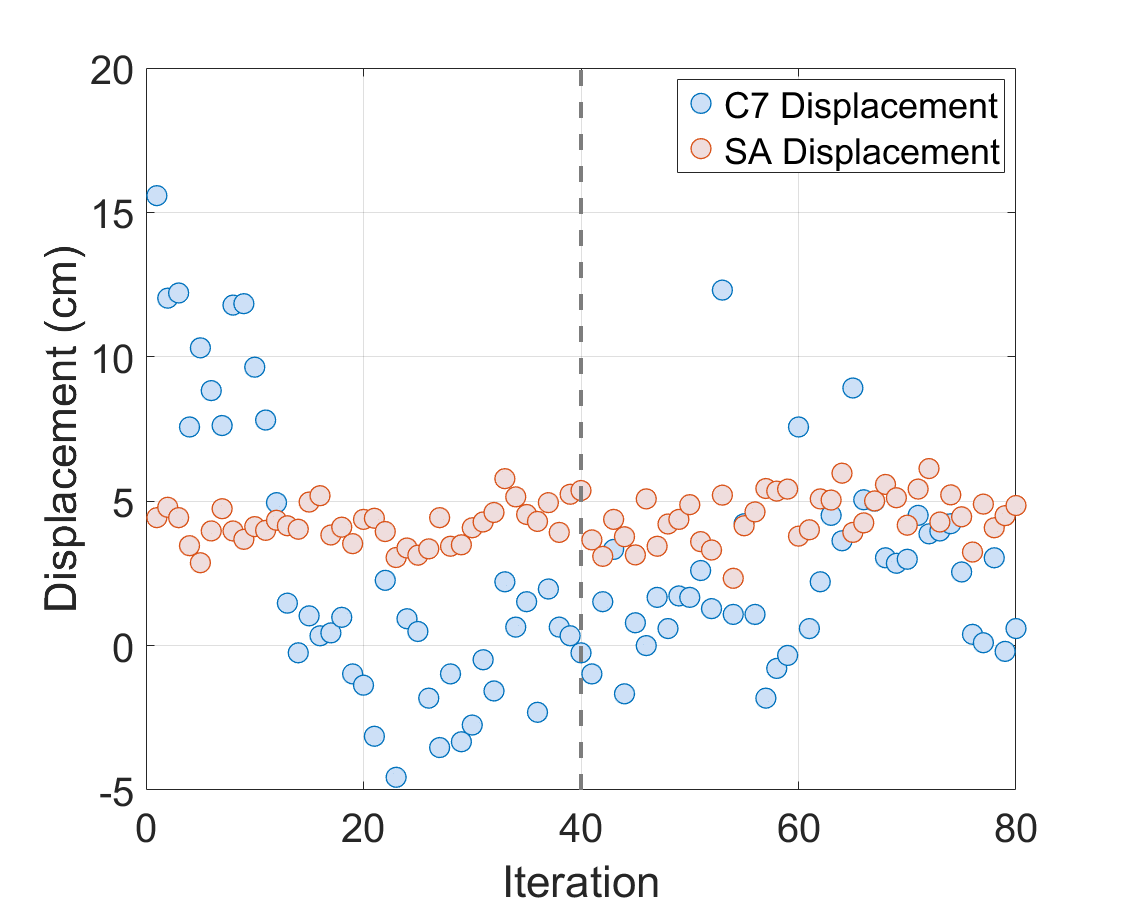}
        \caption{Subject 1.}
        \label{fig:dispS1}
    \end{subfigure}
    ~
    \begin{subfigure}[t]{0.28\textwidth}
        \centering
        \includegraphics[width=1.0\textwidth]{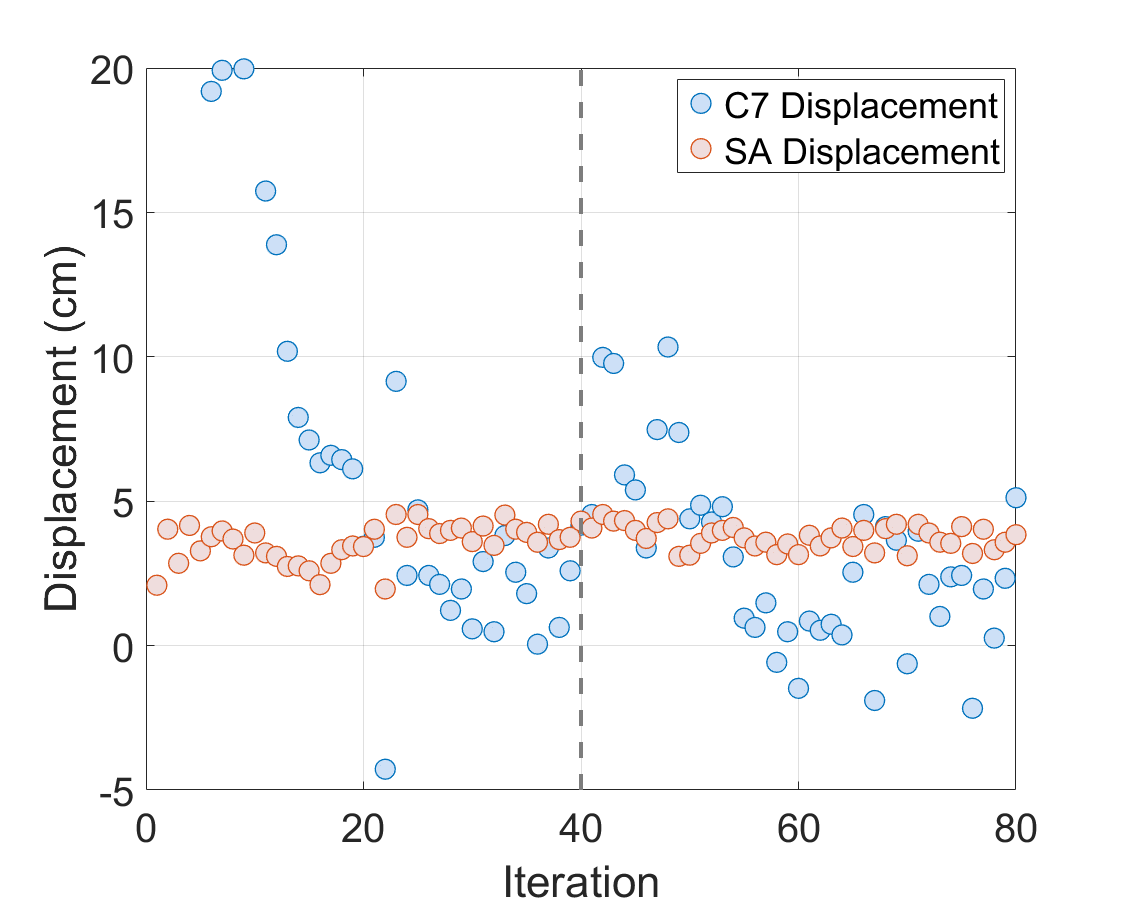}
        \caption{Subject 2.}
        \label{fig:dispS2}
    \end{subfigure}
    ~
    \begin{subfigure}[t]{0.28\textwidth}
        \centering
        \includegraphics[width=1.0\textwidth]{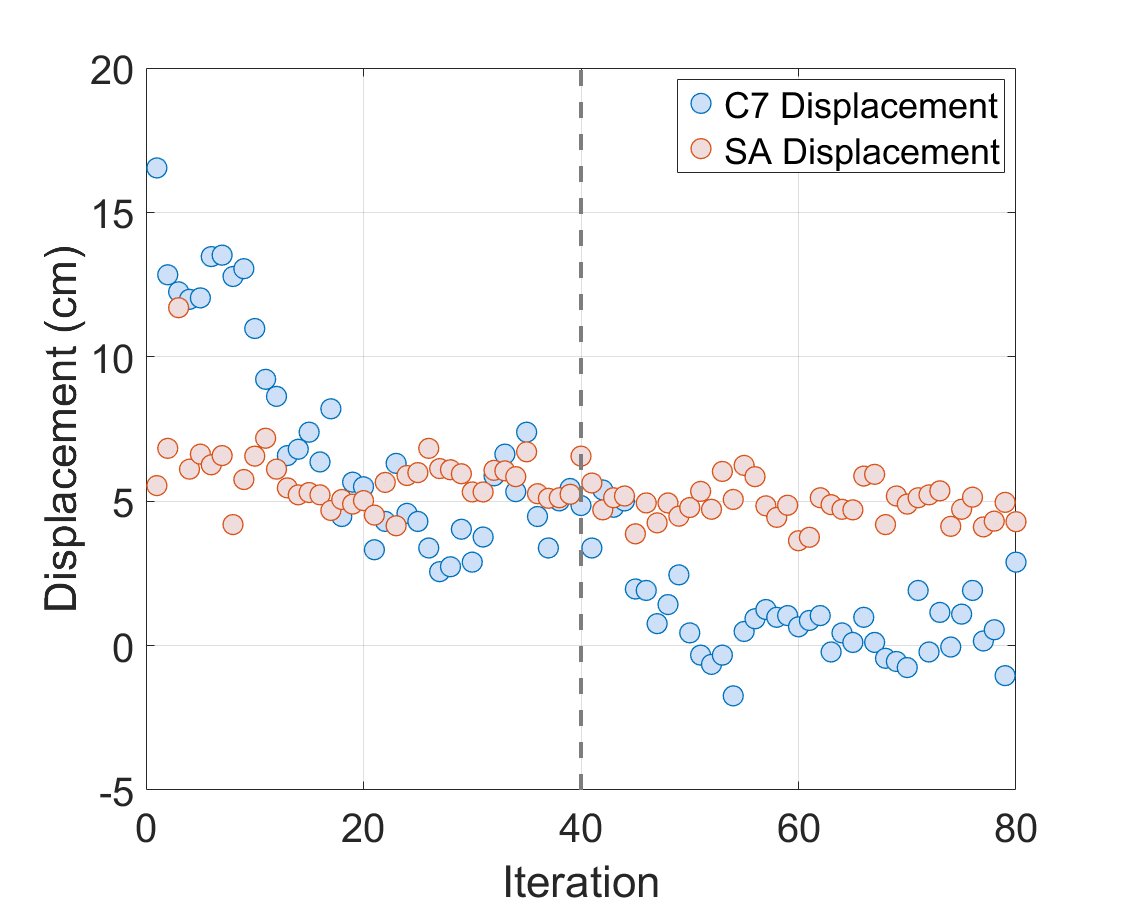}
        \caption{Subject 3.}
        \label{fig:dispS3}
    \end{subfigure}
    
    \begin{subfigure}[t]{0.28\textwidth}
        \centering
        \includegraphics[width=1.0\textwidth]{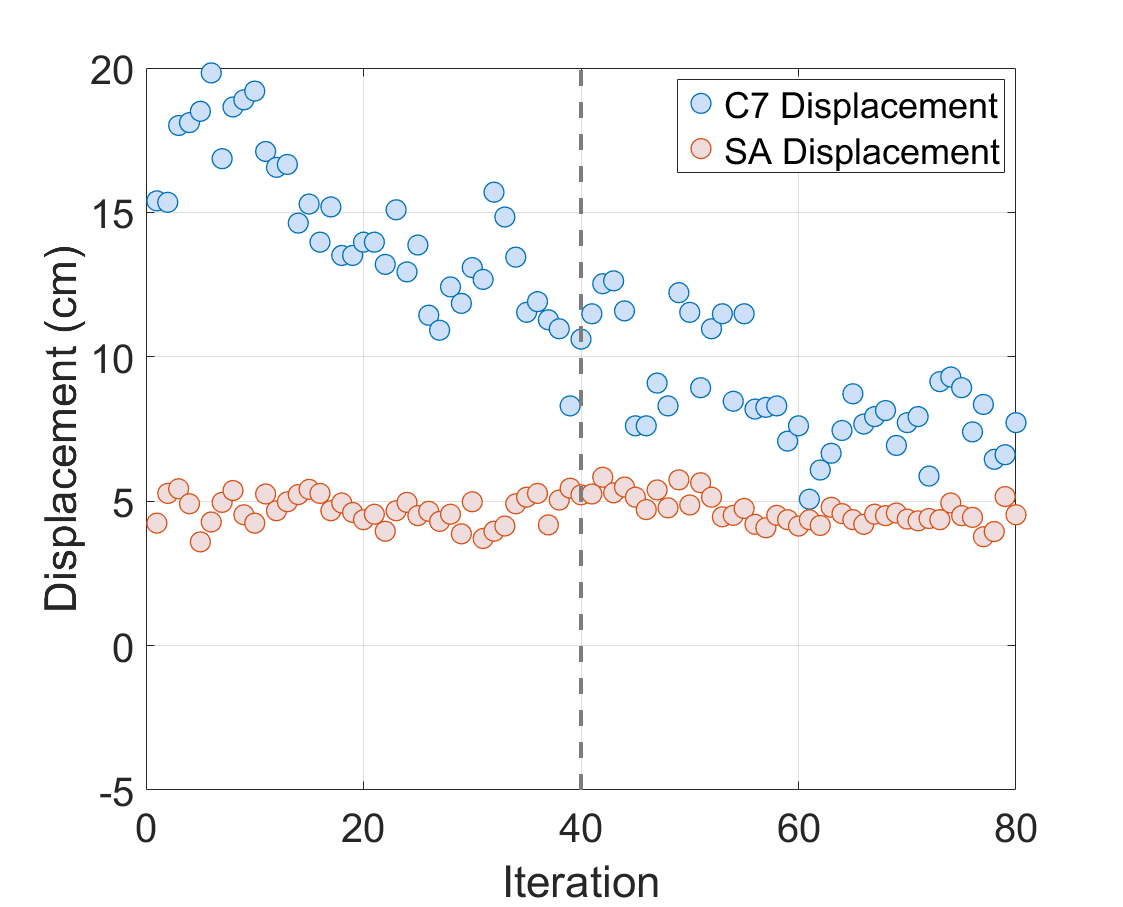}
        \caption{Subject 4.}
        \label{fig:dispS4}
    \end{subfigure}
    ~
    \begin{subfigure}[t]{0.28\textwidth}
        \centering
        \includegraphics[width=1.0\textwidth]{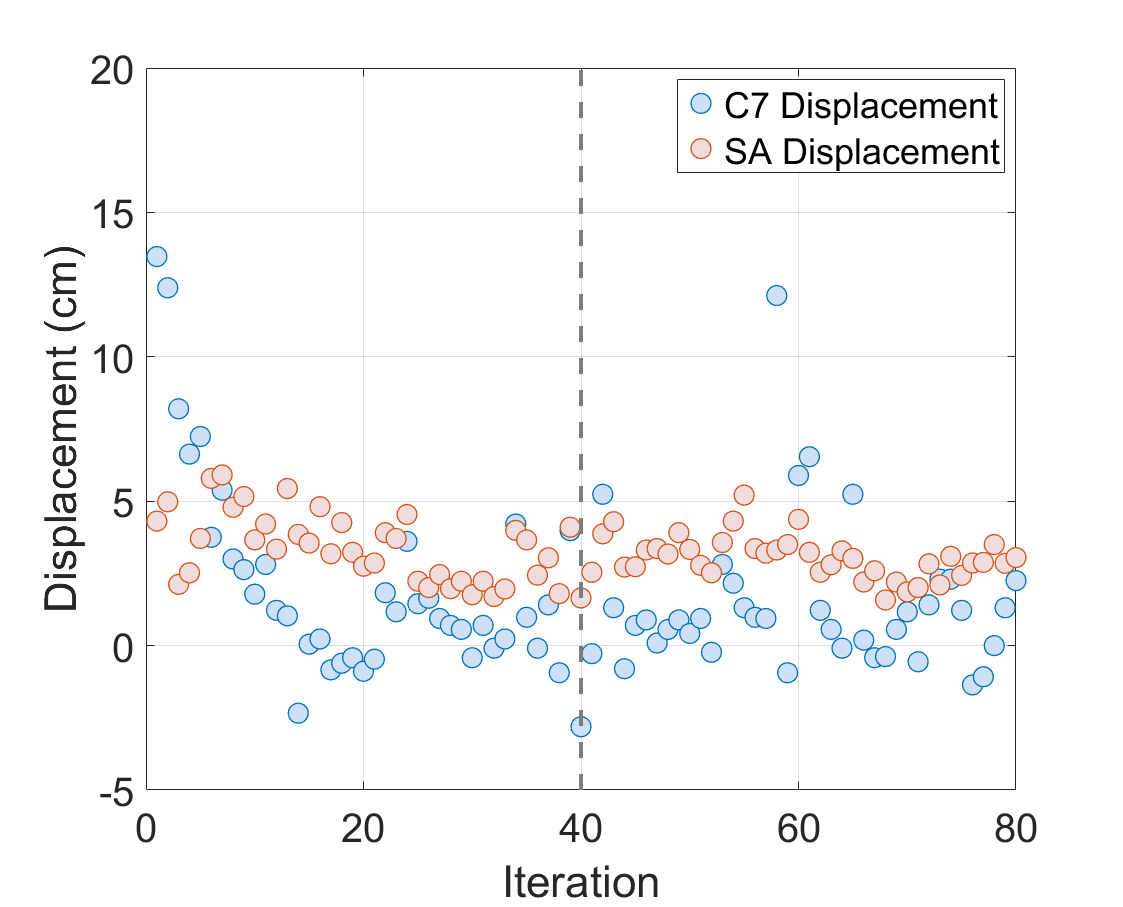}
        \caption{Subject 5.}
        \label{fig:dispS5}
    \end{subfigure}
    ~
    \begin{subfigure}[t]{0.28\textwidth}
        \centering
        \includegraphics[width=1.0\textwidth]{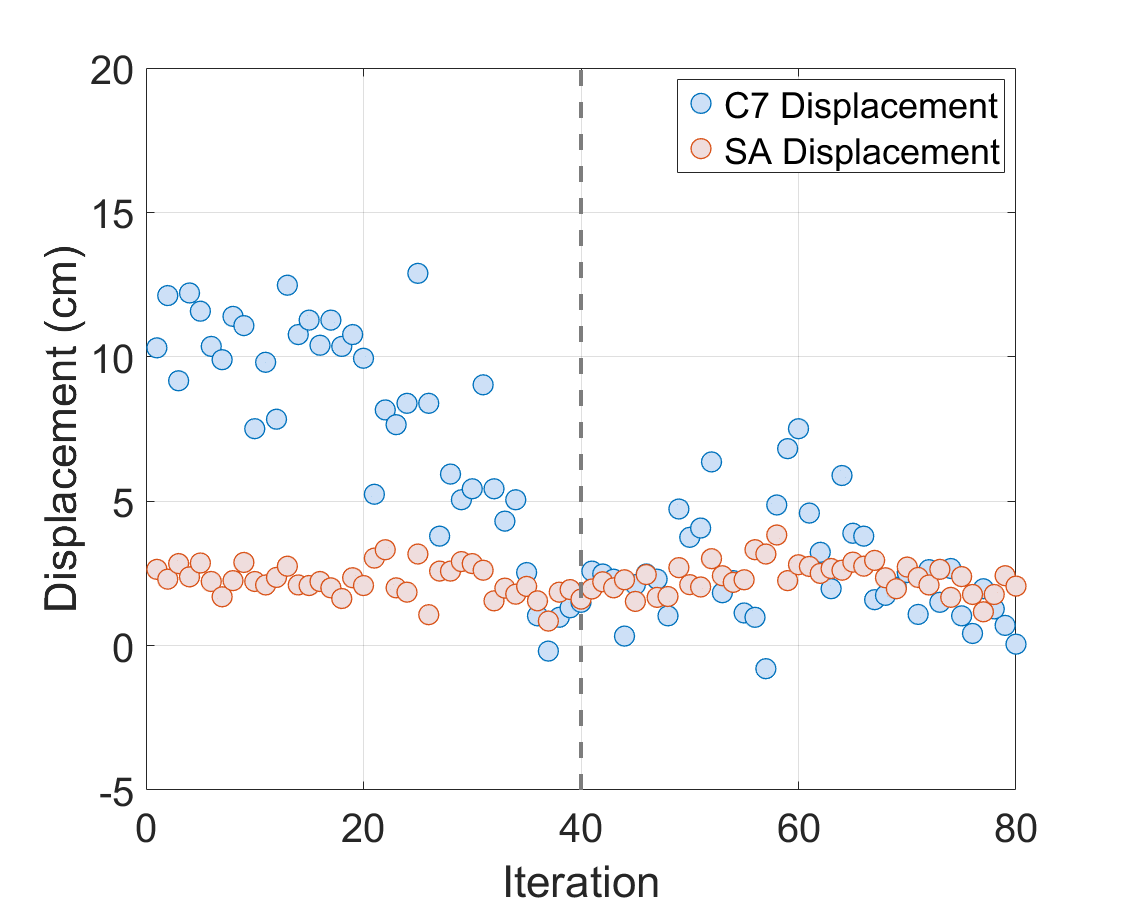}
        \caption{Subject 6.}
        \label{fig:dispS6}
    \end{subfigure}
    
    \begin{subfigure}[t]{0.28\textwidth}
        \centering
        \includegraphics[width=1.0\textwidth]{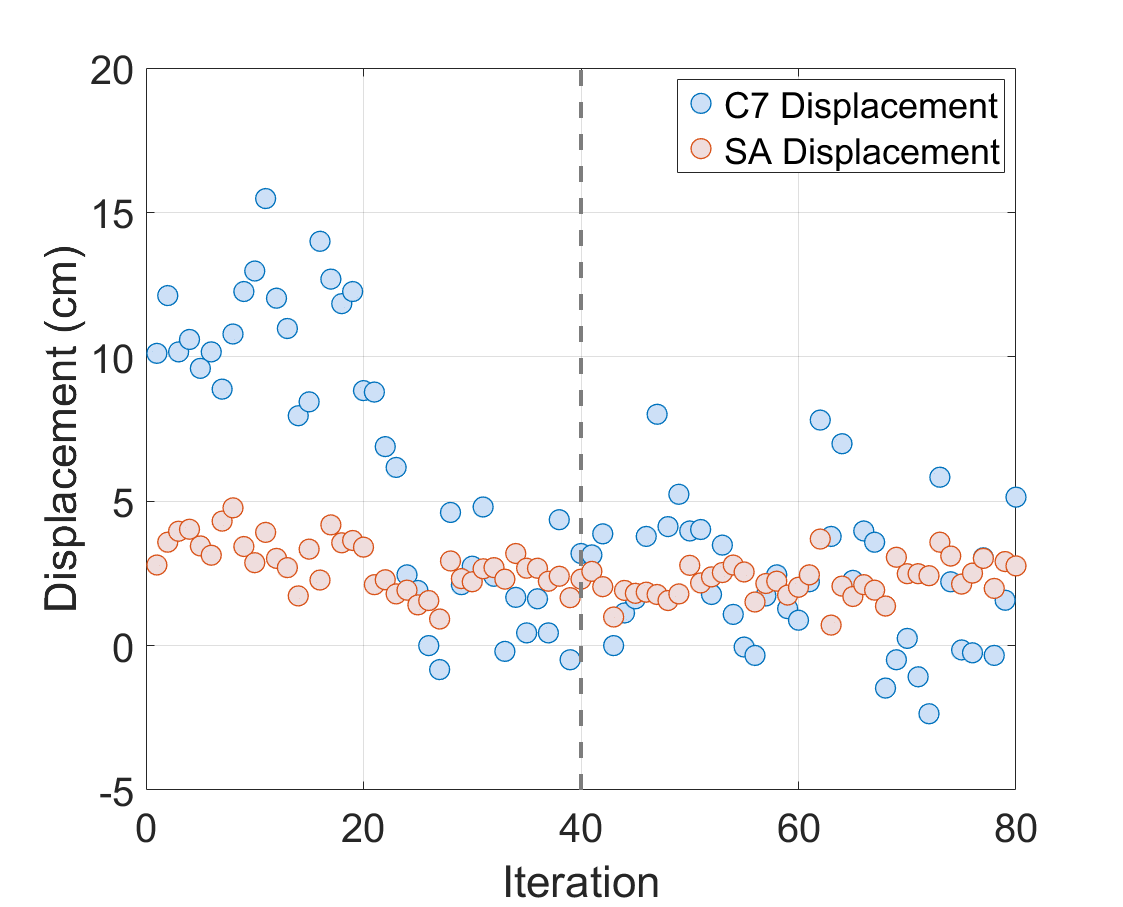}
        \caption{Subject 7.}
        \label{fig:dispS7}
    \end{subfigure}
    ~
    \begin{subfigure}[t]{0.28\textwidth}
        \centering
        \includegraphics[width=1.0\textwidth]{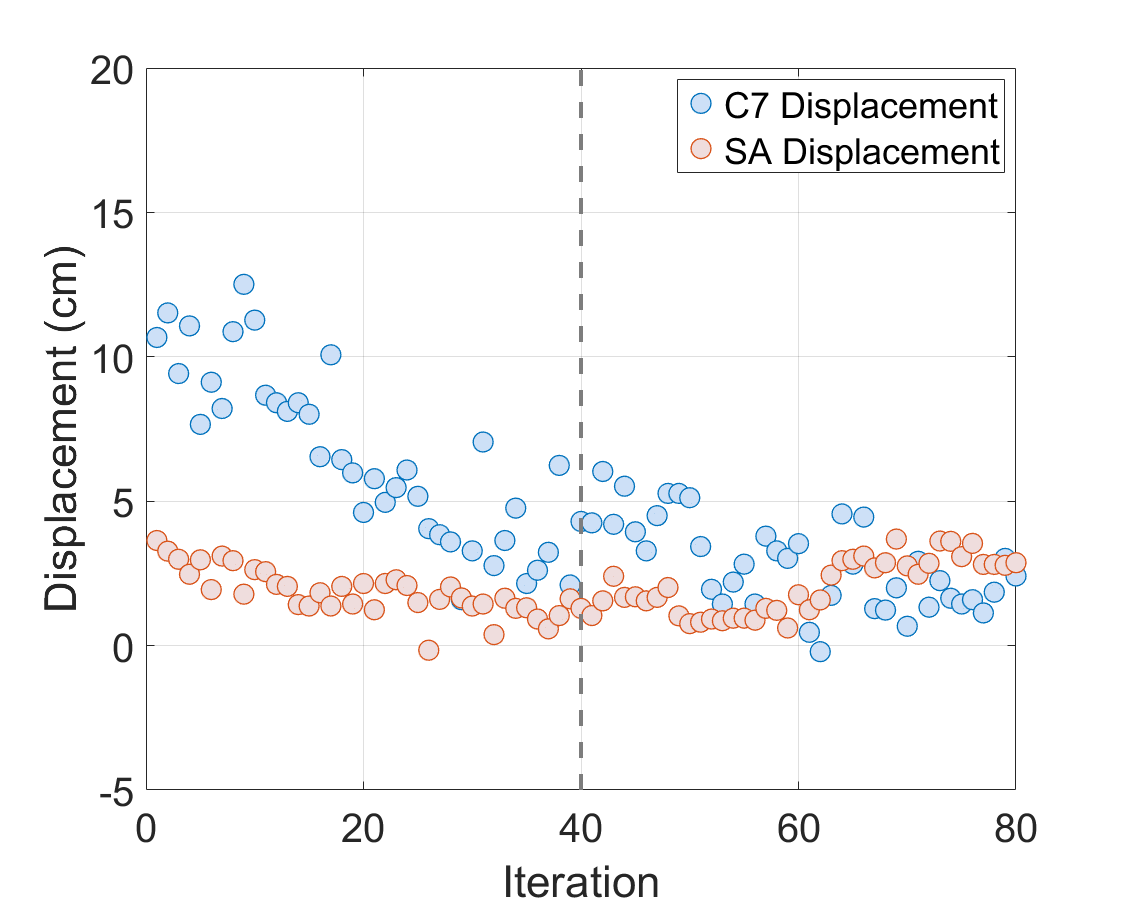}
        \caption{Subject 8.}
        \label{fig:dispS8}
    \end{subfigure}
    ~
    \begin{subfigure}[t]{0.28\textwidth}
        \centering
        \includegraphics[width=1.0\textwidth]{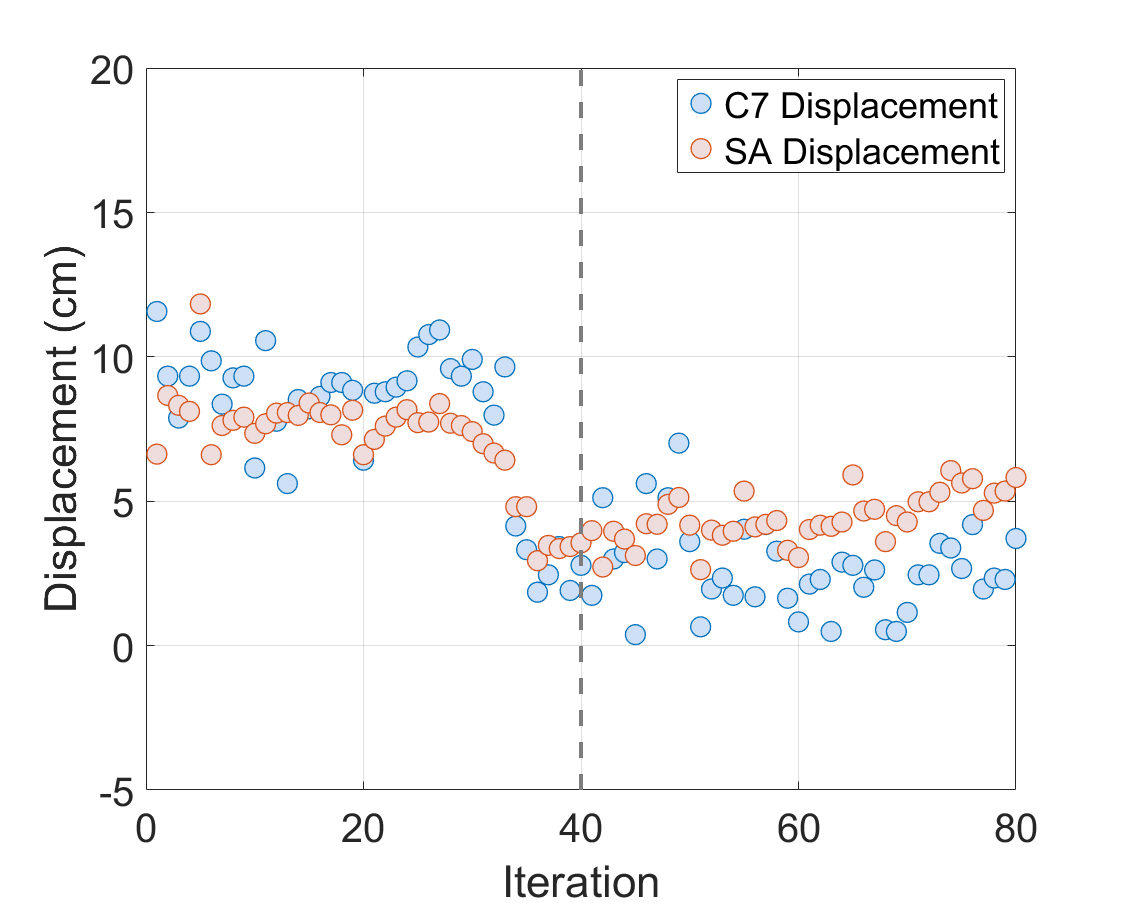}
        \caption{Subject 9.}
        \label{fig:dispS9}
    \end{subfigure}
    \caption{Trunk (C7) and shoulder (SA) displacement over iterations results for nine able-bodied subjects. The blue circles show trunk displacement while the red circles show shoulder displacement. The grey dotted line represents when the subjects were given rest time.}
    \label{fig:displacementResults}
\end{figure*}

Figure \ref{fig:displacementResults} presents trunk (C7) and shoulder (SA) displacement over the 80 iterations of the task. It can be observed that all subjects reduced their trunk motion and achieved under 5cm trunk displacement at the steady-state, except for subject 4 which did not reach steady-state. This steady-state trunk behaviour is comparable to the able-bodied behaviour presented in Figure \ref{fig:synergyisplacementMap}. On the other hand, subjects maintained their ``natural'' shoulder displacement regardless of the objective function using a desired zero shoulder displacement. This can be seen by the constant shoulder displacement throughout the experiment. Only in the results for subject 9, Figure \ref{fig:dispS9}, a change in shoulder strategy can be observed. Nevertheless, these results demonstrate that the personalisation algorithm successfully achieved its objective of minimising compensation motion without providing explicit feedback to its user, for a variety of individuals with different motor strategies. This supports the hypothesis that a surrogate of implicit human motor behaviour can be used as the objective function to drive prosthesis personalisation without the need to provide the prosthesis user with explicit performance feedback. The next section presents a discussion of the important observations arising from these results.

% Results

% Results

%% file: 5_discussion.tex
\section{Discussion}

\subsection{Non-linearities in Human Motor Learning}
The results from subjects 2 and 9, Figures \ref{fig:thetaJS2}, \ref{fig:thetaJS9}, \ref{fig:dispS2}, and \ref{fig:dispS9}, highlight some of the non-linear phenomena present in human motor learning which directly affect the performance of the personalisation algorithm. In the case of subject 2, one of such non-linear phenomena can be observed after the rest period. Here, the subject's cost saw a sudden increase (increased compensation), which was gradually reduced until reaching the steady-state again. This behaviour change can be considered as a ``warm-up decrement'', which is commonly observed at the beginning of experiment sessions that are performed days apart \cite{Rosenbaum2009, Newell2001}. This phenomenon is characterised by a decrease in performance between sessions, followed by a systematic increase in performance which satisfies the monotonic increase to a steady-state condition required by the algorithm. However, in the case of subject 2, it is present within a session as rest time was provided halfway through the experiment. This highlights individual variation in motor learning, which in the case of subject 2 this particular aspect was more prevalent. As can be seen in Figure \ref{fig:dispS2}, these results suggests that within the rest time the subject ``forgot'' the steady-state upper-body and arm motion, and had to re-learn it. This phenomenon could be prevalent in day-to-day prosthesis use and thus is relevant for the further development and subsequent deployment of adaptive systems in human applications.

In the case of subject 9, Figures \ref{fig:thetaJS9} and \ref{fig:dispS9}, there is a sudden change in performance around iteration 35, where the cost ($J_p$) was significantly reduced from one iteration to another. This corresponds to a change in strategy that led to a reduction in trunk displacement, as seen in Figure \ref{fig:dispS9}. The change of strategy can be attributed to natural motor variability and exploration, which can lead to the discovery of new strategies that result in improved performance \cite{Rosenbaum2009, Wilson2014}. This highlights a challenge of using a surrogate of implicit human behaviour as the measure of performance for the personalisation algorithm. The use of incomplete information on the human internal objective function, which relies on certain assumptions, does not fully capture the whole human motor and learning behaviour. This can lead to undesirable phenomena in the measure of performance as observed. In this instance, the discovery of the new strategy led the subject to a new motion steady-state with reduced upper-body compensation, which did not affect the performance of the personalisation algorithm. However, the effects of reduced information on the measure of performance, and motor strategy changes, ought to be thoroughly investigated as these will naturally occur in daily prosthesis use. 

\subsection{Effects of the Implicit Objective Function on Human Motor Behaviour and Prosthesis Personalisation}
There are two observations on the influence of the proposed objective function on motor behaviour that warrant discussion. The first observation is related to compensation (cost) minimisation being achieved for a range of synergy values. As shown in the identified synergy-cost maps in Figure \ref{fig:synergyPerformanceMaps}, minimum cost is observed in a range of synergy values. Results in Figure \ref{fig:synergyPerformanceResults} for subjects 1-3, show how the algorithm converges to a value within this range (for subject 2, iterations 40-50 as per the previous discussion). This is as expected as the algorithm's objective is cost minimisation. However, the possible reasons behind why the map has such flat regions instead of a more convex shape as observed in \cite{Garcia-Rosas2019} warrants discussion. The authors identified two possible causes for such behaviour. 1) There may be a synergy range where the resultant compensation motion is within the subject's natural variation, and thus any synergy on that range may be suitable for the subject. 2) The flat region could be a result of a loss of information due to the use of only trunk and shoulder displacement, and wearable sensors. However, given the results obtained for this study, it is not possible to conclude which was the cause. This will be considered in future work as testing these possibilities may require longitudinal studies that include the change of the synergy-performance map over multiple sessions and time. This includes determining the effects of long-term motor learning on the synergy-performance map and the effects of information uncertainty and loss on the personalisation procedure.

The second observation is related to the influence of the prosthesis objective function on natural human motor behaviour. The use of a non-natural desired shoulder motion ($\bar{x}_{s}=0$) in the cost function did not influence the resultant human motor behaviour. This can be seen in Figure \ref{fig:displacementResults}, where most subjects maintained their natural shoulder motion (SA displacement of $5cm$) throughout the experiment regardless of the synergy value. A possible explanation for this is that the human internal objective function penalises shoulder more heavily than trunk for this type of motion, thus most subjects prefer to recruit their trunk to compensate for the inaccurate synergy. In subjects 8 and 9 (Figures \ref{fig:dispS8} and \ref{fig:dispS9}) a clear change of shoulder displacement was observed. Subject 8 gradually reduced shoulder displacement to close to zero displacement until iteration 60, where a gradual increase started that eventually converged to about $4cm$, close to the population average natural shoulder displacement ($5cm$). This happened around the same iteration that the subject's trunk displacement reached the vicinity of natural trunk motion. On the other hand, subject 9 had a change of strategy around iteration 35, where both trunk and shoulder displacements changed from compensatory-like to near natural. These results suggest that regardless of the implicit cost function used in the prosthesis, individuals will converge to their natural motor behaviour when the synergy is within the range that allows for it. Similar results have been observed in rehabilitation applications \cite{Fong2019}. This agrees with the previous discussion point, suggesting that there is a range of synergy values that achieve minimum compensation. This may be advantageous from a practical implementation perspective as it may be possible to simplify the personalisation procedure and sensor requirements.

While there are important observations that arise from the results in the study presented herein, the successful implementation of the prosthesis personalisation algorithm under practical constraints using a prosthesis objective function unknown to the human proves the hypothesis considered in this work. These results motivate further studies to better understand the interaction between the human and the adaptive prosthesis.

%% file: 6_conclusion.tex
%
% Conclusion
%
\section{Conclusion}
This paper presented a method that uses implicit human motor behaviour as the basis for the personalisation of human-prosthetic interfaces. The method allows the personalisation algorithm proposed by the authors in \cite{Garcia-Rosas2019} to operate without the need to provide explicit performance feedback to the human and using only wearable sensors. A measure of performance that acts as a surrogate for internal human measures of motor performance was used, ensuring that both the human and prosthesis have the same objective. A hardware implementation was performed where the prosthesis measure of performance was based on compensation motion gathered from wearable IMUs. These initial results are promising and indicate using this surrogate approach can lead to successful interface personalisation with reduced compensatory motion in individuals. Further studies are required to consider the minimal sensing requirements in the wearable IMUs to achieve a satisfactory level of personalisation.